\documentclass[10pt,twocolumn,letterpaper]{article}

\usepackage{cvpr}              

\usepackage[accsupp]{axessibility}

\usepackage{times}
\usepackage{epsfig}
\usepackage{graphicx}
\usepackage{amsmath}
\usepackage{amssymb}
\usepackage{booktabs}
\usepackage{bbding}
\usepackage{cuted}
\usepackage{subcaption}
\usepackage{multirow}

\usepackage[pagebackref,breaklinks,colorlinks]{hyperref}

\usepackage[capitalize]{cleveref}
\crefname{section}{Sec.}{Secs.}
\Crefname{section}{Section}{Sections}
\Crefname{table}{Table}{Tables}
\crefname{table}{Tab.}{Tabs.}

\begin{document}

\title{Reference-based Motion Blur Removal\\Learning to Utilize Sharpness in the Reference Image}

\author{
Han Zou$^{1,2}$
~~~~Masanori Suganuma$^{1,2}$ 
~~~~Takayuki Okatani$^{1,2}$
\\
$^{1}$Graduate School of Information Sciences, Tohoku University
~~~~$^{2}$RIKEN Center for AIP \\
{\tt\small \{hzou, suganuma, okatani\}@vision.is.tohoku.ac.jp}
}

\maketitle

\begin{abstract}
Despite the recent advancement in the study of removing motion blur in an image, it is still hard to deal with strong blurs. While there are limits in removing blurs from a single image, it has more potential to use multiple images, e.g., using an additional image as a reference to deblur a blurry image. A typical setting is deburring an image using a nearby sharp image(s) in a video sequence, as in the studies of video deblurring. This paper proposes a better method to use the information present in a reference image. The method does not need a strong assumption on the reference image. We can utilize an alternative shot of the identical scene, just like in video deblurring, or we can even employ a distinct image from another scene. Our method first matches local patches of the target and reference images and then fuses their features to estimate a sharp image. We employ a patch-based feature matching strategy to solve the difficult problem of matching the blurry image with the sharp reference. Our method can be integrated into pre-existing networks designed for single image deblurring. The experimental results show the effectiveness of the proposed method.  
\end{abstract}

\section{Introduction}
\label{sec:introduction}
The removal of motion blurs in an image is one of the fundamental problems of image restoration. Researchers have considered problems in several different settings: whether the blur kernel is known or unknown; whether the blur kernel is spatially constant or varying in the input image; and whether the input is a single or multiple image. 

The employment of deep learning has led to great success even with the most challenging setting, i.e., single-image deblurring in the case of an unknown, spatially varying blur kernel. However, there is a limit for the single-image methods, since it is hard to restore information lost due to a large motion blur. There is the same trade-off as super-resolution, i.e., the trade-off between the naturalness of output images and their precision (i.e., the error from the ground truths)\cite{blau2018perception}; an excessive attempt will lead to ``hallucination,'' i.e., the generation of fake image textures. 

A promising way to overcome this limit is to use multiple images. A typical example is video deblurring, that is, the removal of blur in the image(s) contained in a video using the aid of other images in the same video. The community has considered two problem settings. One is to remove blur in an image in the video by utilizing the contents of adjacent, sharp images(s), assuming their availability. The other is to attempt to recover a ``latent'' image from a sequence of (all) blurry images. In this paper, we consider the problem of removing motion blur in an image of a scene using an additional image as a reference. The reference image is ideally a sharp image of the same scene, and we design our method to work best in that case. However, it works also well when using an image of a different scene as a reference. Our experiments show that the method can utilize a blurry image of the same or even a different scene as a good reference.

It is vital from a practical standpoint to have fewer restrictions on the reference images. Methods that assume the availability of multiple images in the same scene are only applicable to deblurring videos. On the other hand, our method could effectively use an image of a different scene as a reference, which will widen the applicability almost to the level of the single image deblurring methods. Why a different scene image could be used as a reference is because deblurring may be spatially local inference performed in an image. Research on natural image statistics dictates that local image patches of natural images have relatively low degrees of freedom \cite{simoncelli2001natural}; they are constrained in a low-dimensional manifold in the high-dimensional space of local patches. Thus, even if the reference is a different scene image, we can utilize its local patches to deblur the target image, as long as the reference image patches have some similarities with the target image patches. 

We propose a method to utilize this local information from the reference images. Specifically, we first match each local patch of the target image with one of the reference image patches. Their matching is deterministic, while we also use the confidence of each matching in subsequent steps. Now, suppose the most ideal case, in which we are given a sharp reference image of the same scene as the target. In this case, the matching of the patches must be aligned with the correspondences of scene points between the images, similarly to optical flows. However, one is blurry and the other is sharp, in our case, making the matching difficult. 

To cope with this, we employ a multi-scale strategy. Specifically, we deal with the problem coarse-to-fine, i.e., starting from the largest (i.e., coarsest) scale and gradually moving to smaller scales. The underlying idea is that blurs will have a smaller impact on matching patches in a coarser resolution. Going one step further, we attempt to predict the sharp image at each scale and use it for patch matching. Specifically, to match local patches between the target to reference images, we use the predicted sharp image instead of the down-sampled target image itself. This solution further mitigates the above concern since it should be easier to match the sharp image with the reference image, provided that the predicted sharp image is sufficiently accurate. 

Considering the excellent performance of recent methods for single-image deblurring, we choose to extend them to utilize a reference image to further improve their performance. We extend them by enriching the local features of the target blurry image with those of the reference image. Specifically, we first match the local patches of the target and reference images, as mentioned above. We then warp the feature map of the reference image using the matching result, which spatially aligns the feature map with that of the target image. Next, we augment the target image feature map with the warped reference image feature map. Note that the single-image deblurring methods only use the pre-augment features to infer sharp images.  Augmenting them with the reference features will leads to feature enrichment, hopefully leading to a better restoration result. 

We employ a coarse-to-fine strategy for patch matching and feature augmentation. Conveniently, this approach is well aligned with different state-of-the-art methods, i.e., MIMO-UNet~\cite{cho2021rethinking}, NAFNet~\cite{chen2022simple}. We employ them as a base architecture and design several modules that can be integrated into them, which perform the above patch matching and feature enrichment steps. We train these modules and the base network as a whole in an end-to-end manner. Note that the proposed modules can be integrated into any architecture having the same multi-scale, coarse-to-fine approach. Thus, if a better single-image deblurring architecture is developed in the future, our method will theoretically be integrated into it to gain further performance improvement.

\section{Related Work}

\subsection{Image Deblurring}

Image deblurring has been studied for a long time. Deep learning based methods have proved its success in image deblurring. Nah et al.~\cite{nah2017deep} propose a multi-scale architecture based on a coarse-to-fine strategy. They also propose the GOPRO dataset, consisting of pairs of blurry and sharp image sequences of the same scene; they synthesize the blurry sequences by averaging successive sharp frames. Tao et al. adopt~\cite{tao2018scale} adopt a recurrent structure to extract features on different scales and recover a sharp image in a coarse-to-fine manner. Gao et al.~\cite{gao2019dynamic} follow the multi-scale architecture and adopt an encoder-decoder network similar to U-Net. They use DenseBlocks to build their network and propose nested skip connections to learn the higher-order residual. Park et al.~\cite{park2020multi} propose a method that iteratively removes blur through a single UNet. The feature maps from the previous iteration are fed into the encoders of the next iteration to generate sharp results progressively.  Cho et al.~\cite{cho2021rethinking} revisit the recent method based on the coarse-to-fine framework and propose MIMO-UNet (multi-input multi-output U-Net) that deals with multi-scale inputs using a single network.  Chen et al.~\cite{chen2022simple} decompose the SOTA methods and identify the essential components. They propose NAFNet Block based on these components and build a strong simple baseline model. 

\subsection{Reference-based Image Restoration}

Reference-based image restoration uses an extra reference image for better image restoration. The task for which it is the most widely employed is super-resolution (SR). Previous studies have shown the effectiveness of transferring features from a high-resolution reference image and combining its features with low-resolution images. Zhang et al.~\cite{zhang2019image} use Patch Match~\cite{barnes2009patchmatch} for matching and transferring features obtained through a feature extractor based on the VGG network~\cite{simonyan2014very}.  Yang et al.~\cite{yang2020learning}, who adopt attention mechanisms based on feature fusion and further improve their models by integrating features across scales. Aligning the target and reference images is prone to errors. Wang et al.~\cite{wang2021dual} propose an aligned attention method for better fusion of features. The proposed modules can well preserve high-frequency features via spatial alignment operations. In addition to SR, the use of additional reference inputs has been successful in deblurring. Xiang et al.~\cite{xiang2020deep} improve video deblurring performance by learning the sharpness of a reference video. They extract sharp information from reference video and fuse it with an optical flow-based deblurring network to generate better results. Li et al.~\cite{li2022reference} and Li et al.~\cite{li2022deep} adopt feature matching and fusion methods on blurry and sharp reference image pairs. Li et al.~\cite{li2022reference} propose a selective fusion module to guide feature fusion, while Li et al.~\cite{li2022deep} use a rank module to explore and transfer more useful information from the reference. Liu et al.~\cite{liu2023reference} decouples the ref-based deblurring task into a single image
deblurring task and a reference transfer task for better utilizing the reference input.   Although its concept is similar to ours, its method shows considerably lower deblurring performance than the current single-image deblurring methods and, therefore, our method. 

\section{Reference-guided Deblurring} 

The problem we consider here is to estimate a sharp image of the blurry input image $I^{blur}$ of a scene, given an additional reference image $I^{ref}$. Note that $I^{ref}$ may be either a sharper image of the same scene or of a different scene.

\subsection{Outline}
\label{arch}

Instead of designing a whole new network for the problem, we design a module to be integrated into a backbone network, i.e., an existing single-image deblurring network. The backbone network is originally designed to receive $I^{blur}$ of a scene and output its sharp version by itself. Our module is integrated into such a backbone, where it updates the backbone's intermediate feature and sends the resulting feature back to the backbone, aiming at improving the backbone's performance. Although the module has a separate design, we train the integrated model (i.e., the backbone plus our module) in an end-to-end fashion. 

Specifically, our module works as follows. First, an intermediate feature $F^{blur}$ that the backbone extracts from $I^{blur}$ is fed into our module. Next, receiving also $I^{ref}$ and $I^{blur}$, our module compare their local features, extracts necessary information from $I^{ref}$, and fuses it with  $F^{blur}$, yielding an updated intermediate features $F^{fusion}$. Finally, $F^{fusion}$ is fed back to the backbone, where $F^{fusion}$ replaces $F^{blur}$ and is used to estimate the sharp image.

\begin{figure*}[hbt!]
\centering
\includegraphics[width=0.95\textwidth]{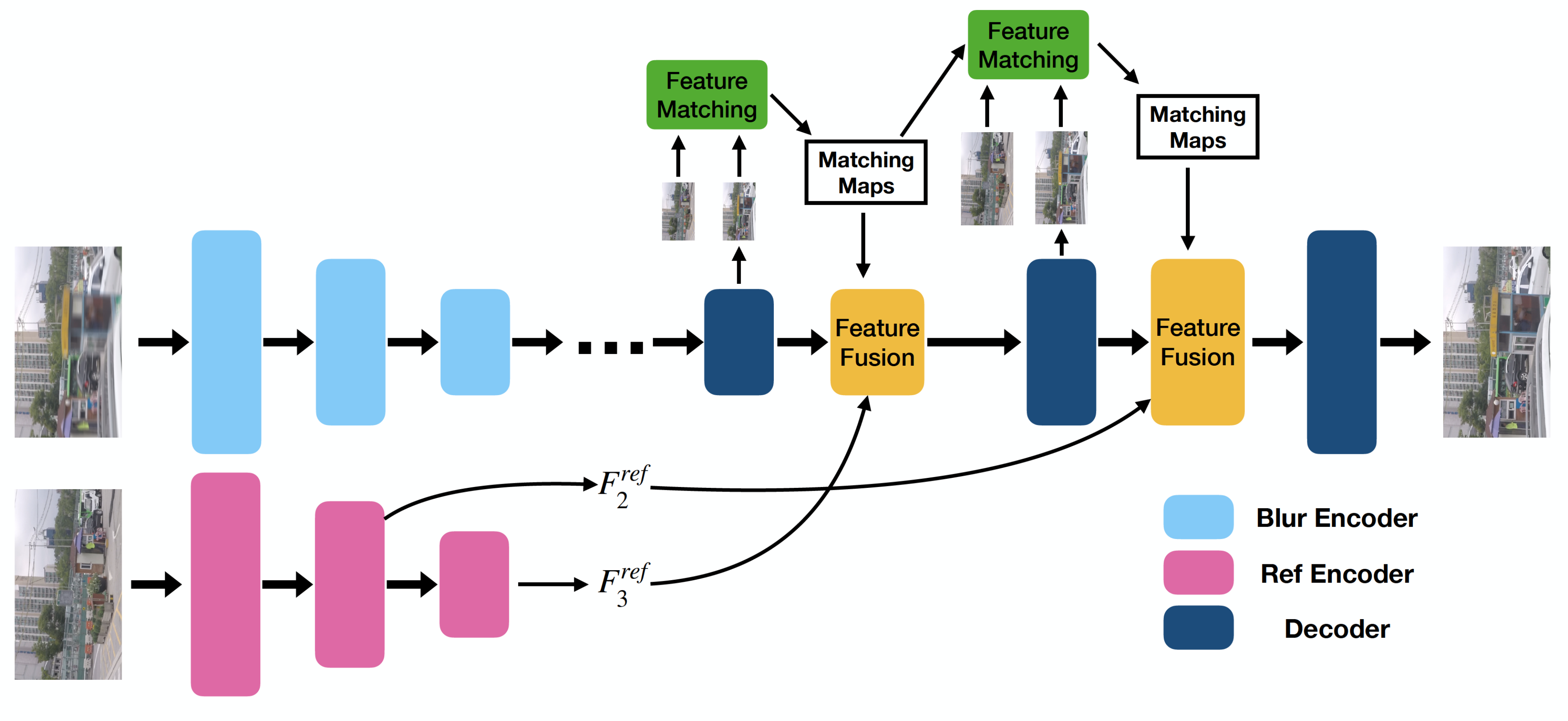}
\caption{Diagram of the proposed network for reference-guided deblurring. It extends UNet architecture. `Feature Matching' and `Feature Fusion' are the newly added components.}
\label{fig:framework}
\end{figure*}

For the backbone network, we primarily consider DeblurNet~\cite{zhou2019davanet}, NAFNet\cite{chen2022simple}, and  MIMO-UNet\cite{cho2021rethinking}, which achieves state-of-the-art performance in the single-image deblurring task. Our module consists of multiple sub-modules that compare/extract/fuse the (features of) blurry and reference images, as above, independently at each scale, as shown in Fig. \ref{fig:framework}. Our method can be used with any backbone network having a similar architecture. 

\subsection{Reference-guided Feature Enrichment}

\subsubsection{Patch Matching on Multi-scale Outputs }

As mentioned above, our module is designed to update the feature $F^{blur}$. The basic idea is as follows. First, comparing $I^{blur}$ (rigorously, the latest estimate of the sharp image) and $I^{ref}$, we find the matching of their local patches, i.e., finding the most similar patch in the latter to each patch of the former (Sec.\ref{sec:match}). Then, by extracting a feature map $F^{ref}$ from $I^{ref}$ using a ref encoder, we spatially divide it into a set of feature vectors and then use the above patch level matches to rearrange them to create a transformed feature map $F^{trans}$. Due to construction, $F^{trans}$ is similar to $F^{blur}$ but differs in that it contains features of a sharper reference $I^{ref}$. Finally, we fuse $F^{trans}$ with $F^{blur}$, replacing $F^{blur}$ in the backbone network (Sec.\ref{sec:fusion}). 

An issue with the above idea is the difficulty in matching (local patches) of $I^{blur}$ and $I^{ref}$ in a meaningful way, since $I^{blur}$ is blurry and $I^{ref}$ is sharp. To cope with this, we use the coarse-to-fine strategy of MIMO-UNet\cite{cho2021rethinking}, i.e., gradually improving estimates from coarser to finer scales while supervising the model to predict the sharp image at each scale. The experimental results in~\cite{cho2021rethinking} show that supervision on outputs of different scales helps to generate sharper intermediate outputs(outputs at low scales) and final outputs.  Inspired by multi-scale supervision, we design a feature matching strategy that matches features between intermediate outputs, denoted by $I^{inter}$, and the reference image $I^{ref}$. $I^{inter}$ has been removed most blurs and tends to have a smaller difference from $I^{ref}$ compared to the downscaled blurry input, making it more accurate to match their local patches and mitigate the above difficulty due to the gap between $I^{blur}$ and $I^{ref}$.

\begin{figure}[htb!]
  \centering
 \includegraphics[width=0.5\textwidth]{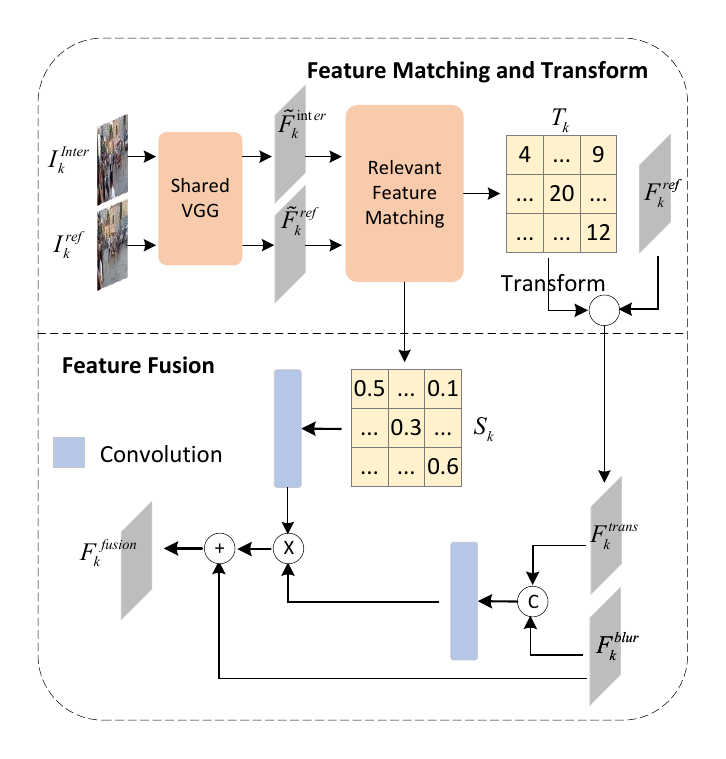}
  \caption{Feature matching and fusion modules used in proposed network. }
\label{fig:fusion}
\end{figure}

\subsubsection{Local Patch Matching of Blurry and Reference Images}
\label{sec:match}

As mentioned above, we employ a coarse-to-fine strategy. We downscale $I^{blur}(\in\mathbb{R}^{H\times W\times 3})$ and $I^{ref}(\in\mathbb{R}^{H\times W\times 3})$ with the factor of $1/2^{k-1}$ $(k=1,\ldots,K)$, obtaining $I^{blur}_k(\in\mathbb{R}^{H_k\times W_k\times 3})$ and $I^{ref}_k(\in\mathbb{R}^{H_k\times W_k\times 3})$, respectively; $H_k=H/2^{k-1}$ and $W_k=W/2^{k-1}$. (Note $I^{blur}_1=I^{blur}$ and $I^{ref}_1=I^{ref}$.)

Starting from the coarsest scale $k=K$, we move from a coarser scale to a finer scale, as in $k=K-1,\ldots,1$. 
At each scale $k$, we obtain an estimate of the sharp image, which we denote by $I^{inter}_k(\in\mathbb{R}^{H_k\times W_k\times 3})$. Note that $I^{inter}_K$ is the final estimate of the sharp image. 

On scale $k$, we calculate the features of $I^{inter}_{k}$ and $I^{ref}_{k}$ that will be used to calculate $I^{inter}_{k-1}$. We first embed $I^{inter}_k$ and $I^{ref}_k$ into feature maps $\tilde{F}^{inter}_k$ and $\tilde{F}^{ref}_k$ using a shared encoder $\phi$. 
Then we extract patches of the size $3\times 3$ from the two feature maps with stride $=1$, yielding $P^{inter}_k=\{p^{inter}_{k,i}\}_{i=1,\ldots,H_k W_k}$ and $P^{ref}_k=\{p^{ref}_{k,i}\}_{i=1,\ldots,H_k W_k}$, respectively.

For matching extracted patches, we calculate the cosine distance $r_{k,i,j}$ between $i$-th element from $P^{inter}_k$ and $j$-th element from $P^{ref}_k$. 

The index $t_i$ of patch most similar to the $i$-th element from $P^{inter}_k$ and its confidence $s_{k,i}$ are given by
\begin{align}
t_{k,i} = \arg \max \limits_{j} r_{k,i,j}, \qquad  s_{k,i}= \max \limits_{j} r_{k,i,j}.
\end{align}

\subsubsection{Feature Fusion}
\label{sec:fusion}

We fuse the features of $I^{ref}_k$ and $I^{blur}_k$ to obtain a feature map that will be used to calculate $I^{inter}_{k-1}$. For the feature map of $I^{blur}_k$, we borrow $F^{blur}_k(\in\mathbb{R}^{H_k\times W_k\times C})$ that is computed for the inference of $I^{inter}_k$ in the backbone deblurring network. For the feature map of $I^{ref}_k$, we compute a new one $F^{ref}_k(\in\mathbb{R}^{H_k\times W_k\times C})$; we input $I^{ref}_k$ into a shallow encoder consisting of 3 stacks of base blocks. 

We then create a new feature map $F^{trans}_k$ from $F^{ref}_k$ using the correspondences between $I^{inter}_k$ and $I^{ref}_k$ represented by $t_{k,i}$ of (\ref{fig:fusion}). To be specific, we generate $F^{trans}_k$ as follows. We denote the spatial coordinates of the feature maps $\tilde{F}^{inter}_k$ and $\tilde{F}^{ref}_k$ by $x$ and $y$; $(x,y)\in [1,W_{k}]\times[1,H_{k}]$. Note that  $\tilde{F}^{inter}_k(x,y)$ and $\tilde{F}^{ref}_k(x,y)\in\mathbb{R}^C$. We then denote the mapping of $(x,y)$ to the patch index by $\mathrm{index}_k(x,y)$ $(:\mathbb{R}^2\mapsto[1,H_{k}\cdot W_{k}])$ and its inverse mapping from an index $i$ to $(x,y)$ by $(x_k(i),y_k(i))$. We then compute $F_k^{trans}(x,y)$ as 

\begin{align}
    F_k^{trans}(x,y) = F_k^{ref}(x_k(t_{k,i}) ,y_k(t_{k,i})),
\end{align}
where $i=\mathrm{index}_k (x,y)$.

We then fuse $F_k^{trans}$ obtained above with $F^{blur}_{k}$. We compute a fused feature map $F_k^{fusion}(\in\mathbb{R}^{H_k\times W_k\times C})$ as follows:
\begin{align}
F^{fusion}_{k} =  \mathrm{conv}_1(F^{blur}_{k}, F^{trans}_{k}) \odot \mathrm{conv}_2(S_k) + F^{blur}_{k},
\label{eqn:fusion}
\end{align}
where  $\mathrm{conv}_1$ and $\mathrm{conv}_2$ are each a single conv layer; $S\in\mathbb{R}^{H_k\times W_k\times 1}$ is a upsampled confidence map; specifically, letting $S_k(x,y)=s_{\mathrm{index}_k(x,y)}$ be the confidence map in the resolution of $H_k\times W_k$. The resulting map $F^{fusion}_{k}$ is upsampled with transposed convolution and fed to the next scale $k+1$. 
In (\ref{eqn:fusion}), our aim is to predict only a `residual' component with the two conv. layers to update $F^{blur}_k$, similarly to the skip connection of ResBlock.

\subsection{Acceleration on Feature Matching}
\label{sec:acceleration}

In order to perform feature matching and fusion across multiple scales, the computational costs can become substantial when dealing with larger scales. To mitigate the computational burden associated with feature matching, we leverage index maps from lower scales and propose a coarse-to-fine approach for acceleration.

\begin{figure}[htb!]
  \centering
 \includegraphics[width=0.4\textwidth]{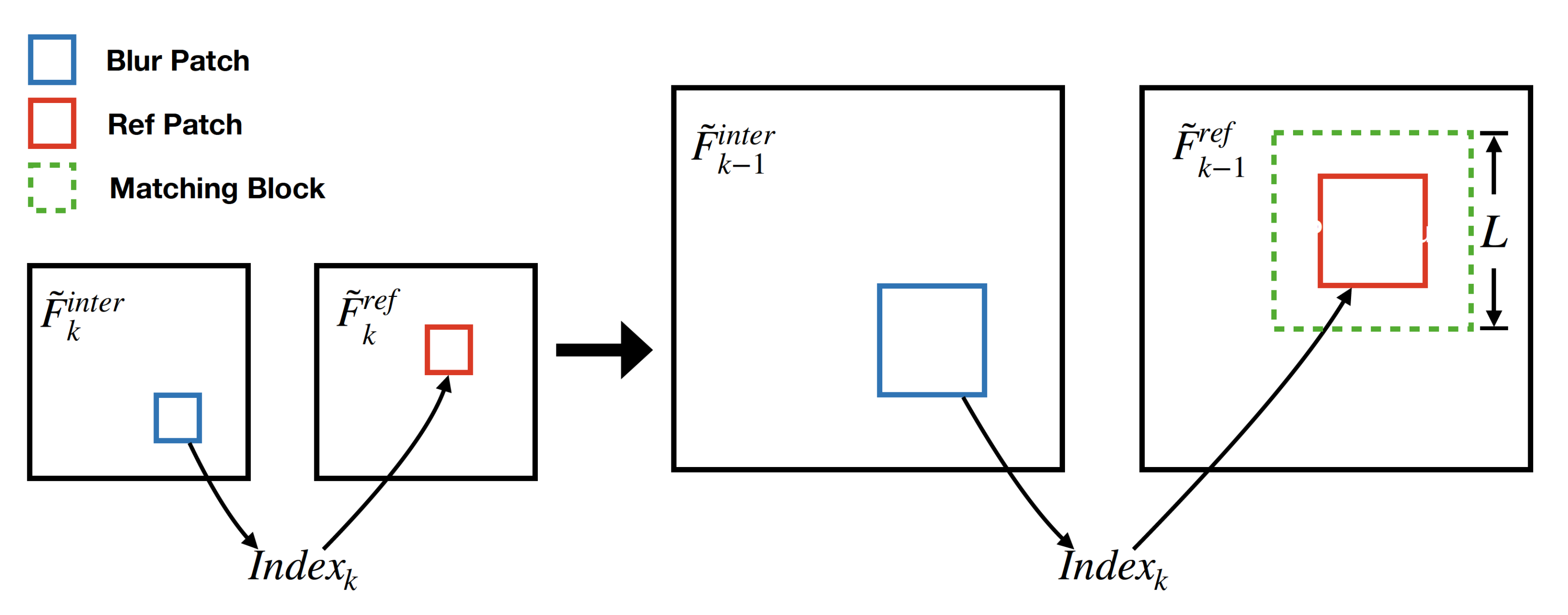}
  \caption{Acceleration method used in the model. The index map obtained from the lower scale, such as $k$, provides an approximate position of the matched patches between the Blur and Ref features. Subsequently, the matching operation at scale $k-1$ can be conducted within a block characterized by a side length of $L$.  }
\label{fig:c2f_match}
\end{figure}

It becomes apparent that images at lower scales contain the same content but with reduced details compared to images at higher scales. This observation suggests that the index map obtained from lower scales can serve as a guide for feature matching at higher scales. By utilizing the coarse index map, there is no longer a necessity for global matching at larger scales. Instead, a localized feature matching around the position indicated by the coarse index map suffices, as shown in the Fig.~\ref{fig:c2f_match}

As discussed in Section~\ref{sec:match}, the computation of cosine distance between all extracted patches $P^{inter}_k=\{p^{inter}_{k,i}\}_{i=1,\ldots,H_k W_k}$ and $P^{ref}_k=\{p^{ref}_{k,i}\}_{i=1,\ldots,H_k W_k}$ is required for patch matching at scale $k$. This process involves $H_k \times W_k \times H_k \times W_k$ operations. However, by utilizing the coarse index map as a guide, the number of operations is reduced to $H_k \times W_k \times L \times L$, where $L$ represents a constant that indicates the side length of a square block within which patch matching is performed.

\subsection{Loss Function}

We train the integrated model, i.e., the backbone network and the proposed feature enrichment module. We follow MIMO-UNet\cite{cho2021rethinking} for the training of the integrated network. It yields an estimate of the sharp image at each scale $k=1,\ldots,K$. For brevity, we denote the sharp image at scale $k$ by $I_k$ and its estimate by $\hat{I}_k=I^{iter}_k$.

We consider the following two losses between $I_k$ and $\hat{I}_k$. The first is the Charbonnier loss defined by
\begin{align}
L_{cb} = \sum \limits^{K}_{k=1} \sqrt{ \|I_{k} - \hat{I}_k\|_{2} + \epsilon^2}.
\end{align}
The second is the frequency reconstruction loss given by
\begin{align}
L_{fr} = \sum \limits^{K}_{k=1} \| \mathcal{F}(I_{k}) - \mathcal{F}(\hat{I}_k)\|_{1}, 
\end{align}
where $\triangle$ and $\mathcal{F}$ denote the Laplacian operator and the fast Fourier transform.
We use the following weighted sum for the total loss for the training:
\begin{align}
L_{total} = \alpha  L_{cb} + \beta L_{fr},
\end{align}
where we set $\alpha$ and $\beta$ to 1 and 0.01, respectively, in our experiments.

\section{Experiments}

\subsection{Experimental Settings}

\subsubsection{Datasets}

We used three datasets in our experiments, GOPRO~\cite{nah2017deep}, RealBlur~\cite{rim2020real}, and HIDE~\cite{shen2019human}. We primarily use the GOPRO\cite{nah2017deep}  datasets for the training of the proposed method. The GOPRO\cite{nah2017deep} dataset contains 2,013 training pairs of 22 different scenes and 1,111 test pairs of 11 different scenes. The RealBlur\cite{rim2020real} dataset contains 3,758 training pairs of 182 different scenes and 980 image pairs of 50 different scenes. It contains two subsets: RealBlur-J\cite{rim2020real}, a set of JPEG images processed by camera ISPs, and RealBlur-R\cite{rim2020real}, those generated from camera raw images. The HIDE~\cite{shen2019human} dataset contains 8,422 pairs of realistic blurry and ground truth images. 

As these datasets do not officially provide reference images, we choose them as follows. For the GOPRO\cite{nah2017deep} dataset, we randomly sample another frame from the same scene for the reference. Specifically, we choose a reference from the range of $[-30,30]$ frames before and after the target frame. We do the same for the RealBlur\cite{rim2020real} and HIDE~\cite{shen2019human} dataset, where we randomly choose one of the same scene images as the target image as its reference. We train our models so that they can utilize not only sharp images but also blurry images, effectively as references. Therefore, we include blurry reference images in the training for both datasets. Specifically, we randomly choose a sharp or a blurry image for each target image; we set their ratio to 8:2 (sharp:blurry). We use the ground-truth sharp images for the sharp references and the input blurry images for the blurry reference at the frames chosen as above. The side length $L$ used in matching acceleration is set to 16. 

\subsubsection{Implementation Details}

We primarily use DeblurNet~\cite{zhou2019davanet}, two variants of MIMO-UNet~\cite{cho2021rethinking} and two variants of NAFNet~\cite{chen2022simple} as backbones.  MIMO-UNet and MIMO-UNet+ employ eight and twenty ResBlocks~\cite{he2016deep} in the encoder and decoder, respectively. NAFNet32 and NAFNet64 employ 32 and 64 channels, respectively. As shown in Fig.~\ref{fig:framework}, we augment it with three components, i.e., feature matching module (Fig.~\ref{fig:fusion}), feature fusion module (Fig.~\ref{fig:fusion}), and an encoder for extracting features from the input reference image. 

The feature matching module first extracts features from $I^{inter}$ (or $I^{blur}$ at the coarsest scale) and $I^{ref}$. We use an ImageNet~\cite{krizhevsky2012imagenet} pretrained VGG19~\cite{simonyan2014very} network for this purpose. To extract features from $I^{ref}$, we use a stack of three convolutional layers followed by four ResBlocks in Ref-DeblurNet and Ref-MIMO-UNet, and use a stack of three convolutional layers followed by 2 NAFBlocks in Ref-NAFNet. 

\subsubsection{Training}

We train all components as a whole in an end-to-end manner using the Adam optimizer~\cite{kingma2014adam} with $\beta_1=0.9$ and $ \beta_2=0.999$. Setting the initial learning rate at $2 \times 10^{-4}$, we employ a learning rate scheduler based on cosine annealing\cite{loshchilov2016sgdr}; the learning rate decreases steadily to $1 \times 10^{-6}$. 

Following previous studies of single-image deblurring, we set the input image size to $256 \times 256$ at the training time. (We input the images with original sizes into the network at the test time.) As the original images have larger sizes, we randomly crop $256 \times 256$ square regions from these images; we crop an identical square from the blurry and its reference images. We apply random horizontal and vertical flipping with a probability $=0.5$ for data augmentation.

\subsection{Experimental Results}

\subsubsection{Quantitative Comparison} 

We first evaluate our proposed method on the GOPRO~\cite{nah2017deep}, HIDE~\cite{shen2019human}, and RealBlur~\cite{rim2020real} dataset. The compared methods are as follows: single image-based methods and reference-based methods. We borrow their results from the respective papers, where a network is trained and tested in the same setting. Note that reference-based methods including ours only need a single frame as additional input.

\setlength{\tabcolsep}{3pt}
\begin{table}[!ht]
\begin{minipage}{0.98\columnwidth}
\caption{Qualitative results on the GOPRO~\cite{nah2017deep} and HIDE~\cite{shen2019human} datasets. Models are trained on GOPRO and tested on GOPRO (3-4th col.) and HIDE (5-6th col.) `Type' column indicates the inputs of model; `S' means a single blurry input; `R' uses an additional reference image together with a blurry input. }
\label{table:comparison}
\begin{center}
\begin{tabular}{lccccc}
\hline
\multirow{2}{*}{Methods} & \multirow{2}{*}{Type} &\multicolumn{2}{c}{GOPRO~\cite{nah2017deep}}   &\multicolumn{2}{c}{HIDE~\cite{shen2019human}}\\\cline{3-6}
&  & PSNR & SSIM & PSNR & SSIM\\
\hline
DeblurNet~\cite{zhou2019davanet}     & S & 30.55 & 0.940 & - & - \\
DeblurGANv2~\cite{kupyn2019deblurgan}& S & 29.55 & 0.934& 26.61 & 0.875 \\
SRN~\cite{tao2018scale}              & S & 30.26 & 0.934 & 28.36 & 0.915 \\
DMPHN~\cite{zhang2019deep}           & S & 31.20 & 0.945 & -     & -     \\
MPRNet~\cite{zamir2021multi}         & S & 32.66 & 0.959 & 30.96 & 0.939 \\
RADN~\cite{purohit2020region}        & S & 31.76 & 0.953 &  -    & -     \\
MIMO-UNet~\cite{cho2021rethinking}   & S & 31.73 & 0.951 & 29.28 & 0.921  \\
MIMO-UNet+~\cite{cho2021rethinking}  & S & 32.45 & 0.957 & 29.99 & 0.930 \\
MIMO-UNet++~\cite{cho2021rethinking} & S & 32.68 & 0.959 & -     &     - \\
HINet~\cite{chen2021hinet}           & S & 32.77 & 0.959 & 30.32 & 0.932 \\
NAFNet32~\cite{chen2022simple}           & S & 32.85 & 0.959 & 30.60 & 0.936 \\
Li et al.~\cite{li2022learning}          & S & 33.28 & 0.964 & - &- \\
NAFNet64~\cite{chen2022simple}           & S & 33.69 & 0.961 & 31.32 & 0.943 \\
\hline 
Li et al.\cite{li2022reference}  & R & 29.73  & 0.902 & -     &     -\\
Li et al.\cite{li2022deep}       & R & 30.31  & 0.900 & -     &     -\\
Liu et al.\cite{liu2023reference} & R & 33.35 & 0.963 & 31.02 & 0.940 \\

\hline 
Ref-DeblurNet   & R & 31.68 & 0.951  & 29.36 & 0.921 \\
Ref-MIMO-UNet  & R & 32.53 & 0.957  & 30.07 & 0.931  \\
Ref-MIMO-UNet+ & R & 33.18  & 0.963  & 30.91 & 0.940  \\
Ref-NAFNet32   & R & 33.22 & 0.964  & 30.96 & 0.940  \\
Ref-NAFNet64   & R & 34.13 & 0.970  & 31.48 & 0.945  \\
\hline
\end{tabular}
\end{center}
\end{minipage}
\end{table}

We train our network on the training set of the GOPRO dataset and test it on the test set of the GOPRO and HIDE datasets. Table \ref{table:comparison} shows the qualitative results. 
We can see that our methods that are integrated into five base models, DeblurNet, MIMO-UNet, MIMO-UNet+, NAFNet32 and NAFNet64. The proposed method show improvements of 1.13dB (30.55 vs. 31.68), 0.8dB (32.53 vs. 31.73),  0.73dB (33.18 vs. 32.45), 0.37dB (33.22 vs. 31.85) and 0.44dB (34.13 vs. 33.69) in PSNR, respectively, on GOPRO \cite{nah2017deep}. 
When testing the same network on HIDE \cite{shen2019human}, we can see that our method yields more improvements, i.e., 0.79dB (30.07dB vs. 29.28dB) and 0.92dB (30.91dB vs. 29.99dB), respectively. 

The comparison results on RealBlur~\cite{rim2020real} are presented in Table \ref{table:rb_comparison}. Our network is trained using the GOPRO dataset and subsequently evaluated directly on RealBlur-R and -J~\cite{rim2020real}. Despite the substantial domain gap between the RealBlur dataset and the GOPRO dataset, the proposed module demonstrates the potential for performance improvement by utilizing a reference image. This improvement is notable considering that the RealBlur dataset consists of non-synthesized blurry images. We have omitted the performance evaluation of DeblurNet and NAFNet64 due to specific reasons. DeblurNet lacks performance data on the RealBlur dataset, while NAFNet64 produces unusual images in certain cases of the RealBlur dataset.

\setlength{\tabcolsep}{3pt}
\begin{table}[!ht]
\begin{minipage}{0.98\columnwidth}
\caption{Results on the RealBlur\cite{rim2020real} dataset.}
\label{table:rb_comparison}
\begin{center}
\begin{tabular}{llllll}
\hline
\multirow{2}{*}{Methods} & \multirow{2}{*}{Type} &\multicolumn{2}{c}{RealBlur-R}   &\multicolumn{2}{c}{RealBlur-J}\\\cline{3-6}
& & PSNR & SSIM & PSNR & SSIM\\
\hline
DeblurGANv2~\cite{kupyn2019deblurgan} & S& 35.26 & 0.944 &28.70 & 0.866 \\
SRN~\cite{tao2018scale}              & S &  35.66 &0.947 &28.56 &0.867 \\
DMPHN~\cite{zhang2019deep}             & S& 35.70 & 0.948 & 28.42 &0.860 \\
MPRNet~\cite{zamir2021multi}          & S& 35.99 & 0.952 & 28.70 & 0.873\\
MIMO-UNet~\cite{cho2021rethinking}   & S &35.47 & 0.946 &27.76 &  0.836 \\
MIMO-UNet+~\cite{cho2021rethinking}   & S &35.54& 0.947& 27.63 & 0.837 \\
NAFNet32~\cite{cho2021rethinking}   & S &35.97 & 0.951 & 28.75 &0.875 \\
\hline
Ref-DeblurNet &R & 35.68  & 0.947  & 28.11  & 0.849  \\
Ref-MIMO-UNet &R & 35.64 & 0.948  &  28.04 & 0.844 \\
Ref-MIMO-UNet+ &R & 35.73 & 0.949  &  28.13 & 0.845 \\
Ref-NAFNet32  &R & 36.13 & 0.956  & 28.91 &  0.879  \\
\hline
\end{tabular}
\end{center}
\end{minipage}
\end{table}

\subsubsection{Qualitative Comparison} 
\begin{figure*}[!ht]
    \centering
     \begin{subfigure}[b]{0.18\textwidth}
         \centering
         \includegraphics[width=\textwidth]{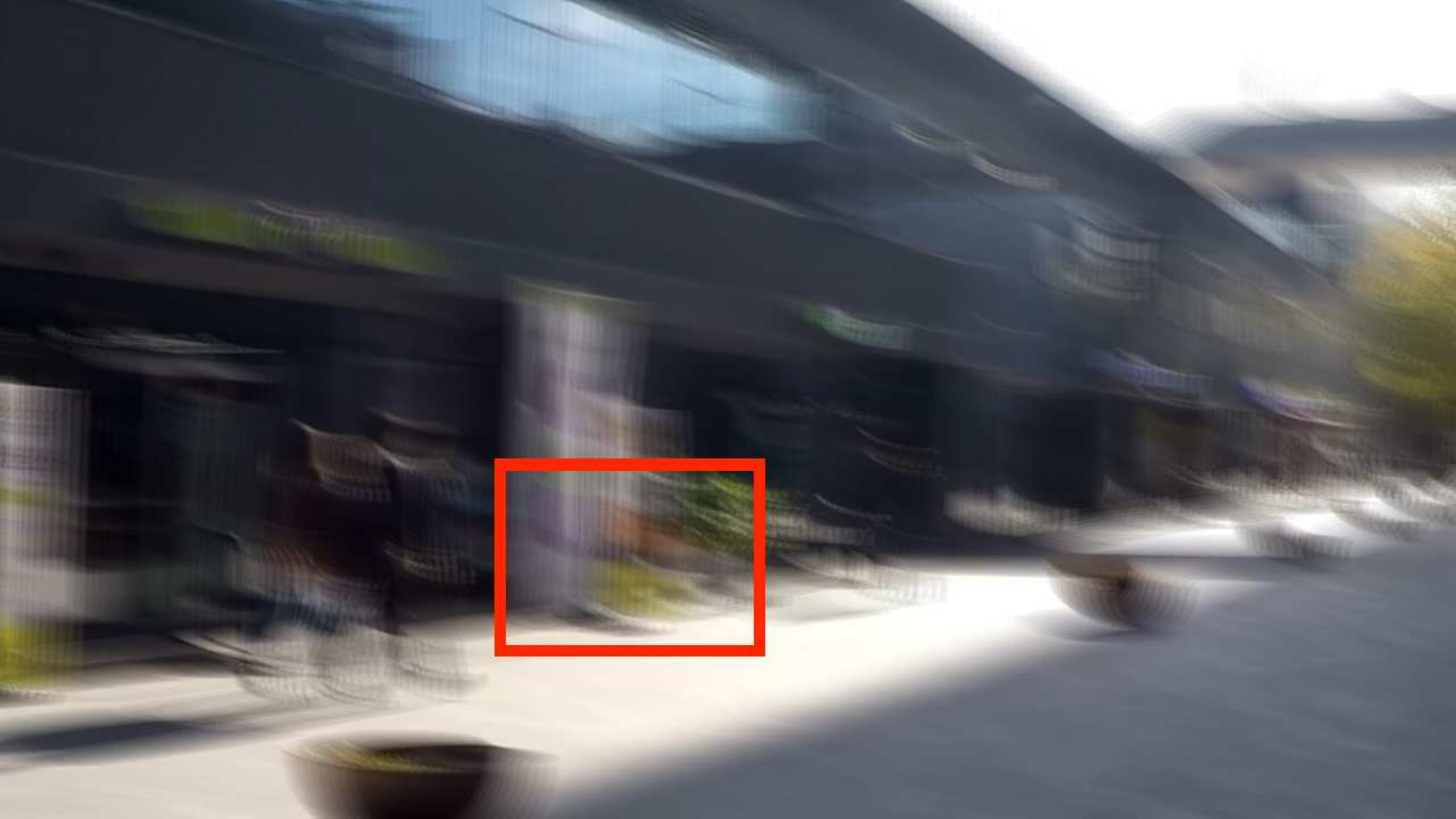}
     \end{subfigure}
     \medskip
     \hfill
     \begin{subfigure}[b]{0.15\textwidth}
         \centering
         \includegraphics[width=\textwidth]{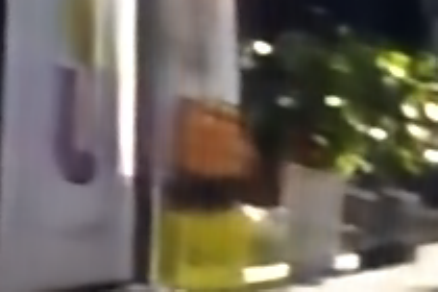}
     \end{subfigure}
     \hfill
     \begin{subfigure}[b]{0.15\textwidth}
         \centering
         \includegraphics[width=\textwidth]{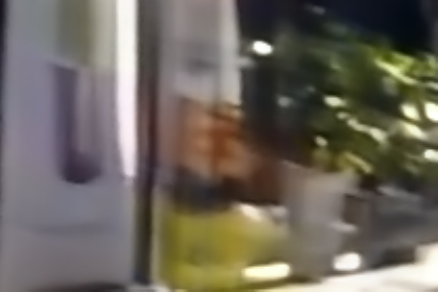}
     \end{subfigure}
     \hfill
     \begin{subfigure}[b]{0.15\textwidth}
         \centering
         \includegraphics[width=\textwidth]{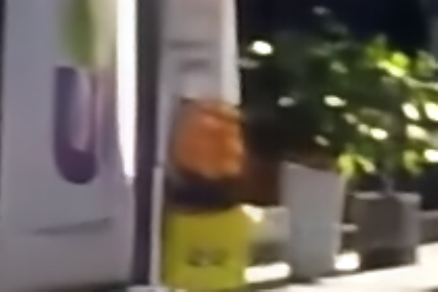}
     \end{subfigure}
     \hfill
     \begin{subfigure}[b]{0.15\textwidth}
         \centering
         \includegraphics[width=\textwidth]{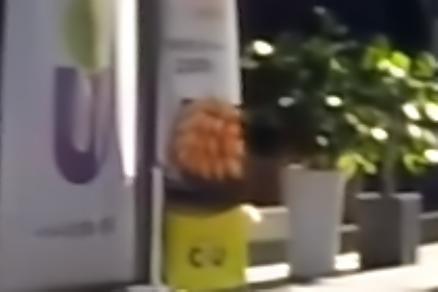}
     \end{subfigure}
     \hfill
     \begin{subfigure}[b]{0.15\textwidth}
         \centering
         \includegraphics[width=\textwidth]{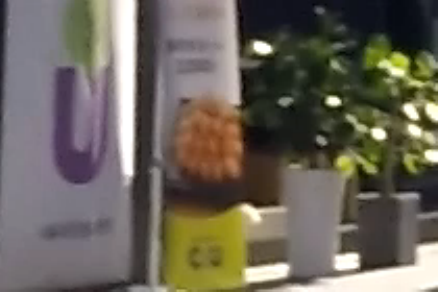}
     \end{subfigure}
     \begin{subfigure}[b]{0.18\textwidth}
         \centering
         \includegraphics[width=\textwidth]{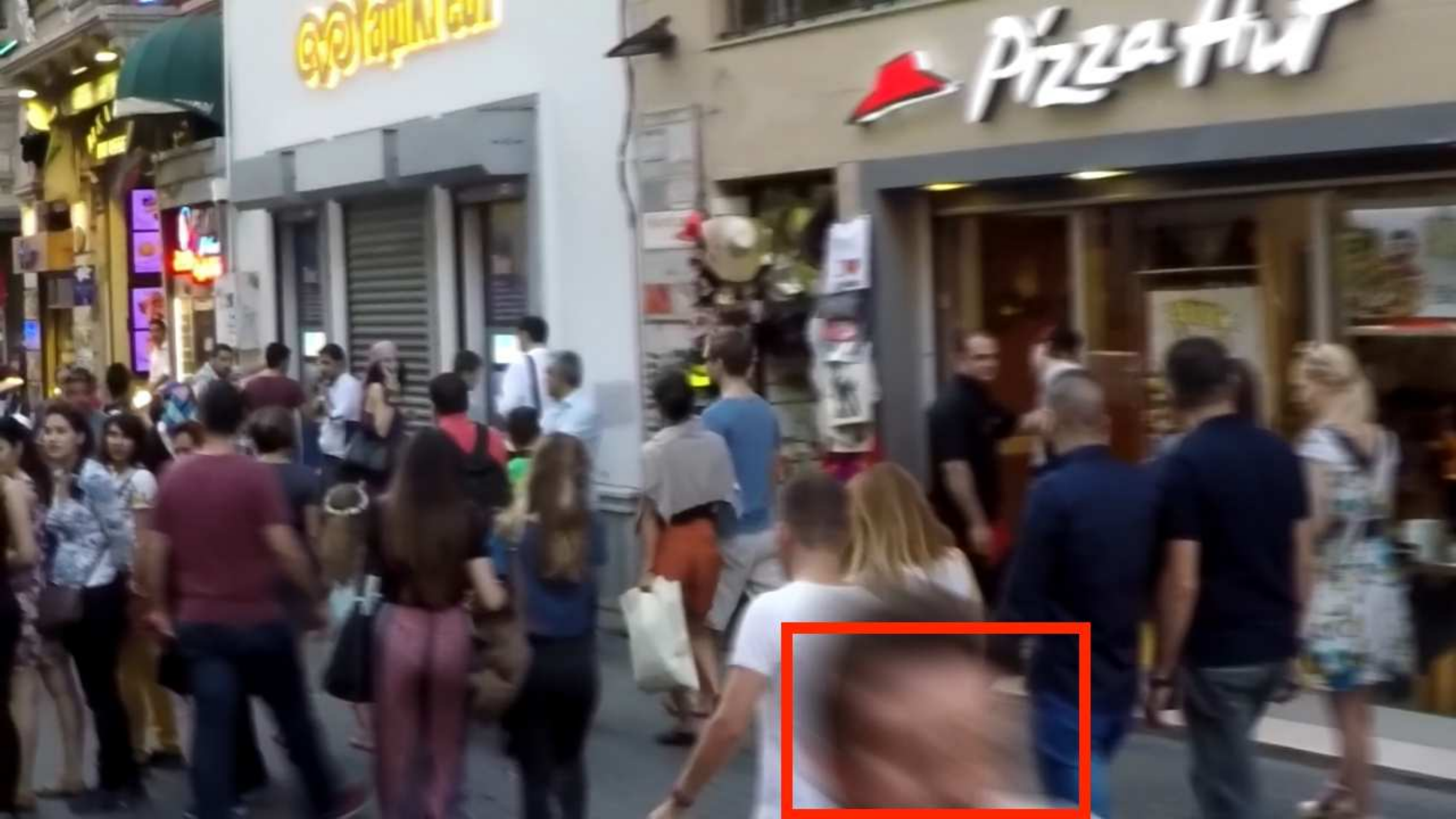}
     \end{subfigure}
     \medskip
     \hfill
     \begin{subfigure}[b]{0.15\textwidth}
         \centering
         \includegraphics[width=\textwidth]{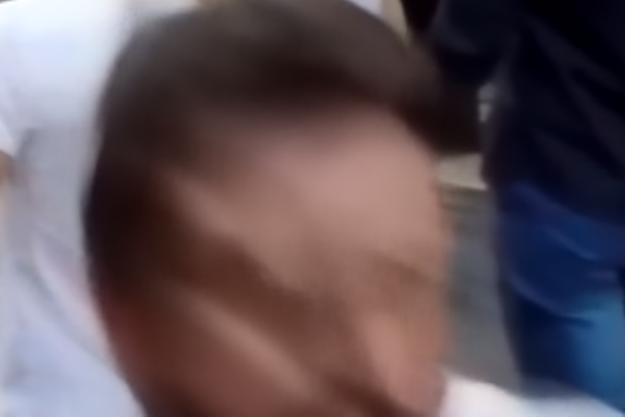}
     \end{subfigure}
     \hfill
     \begin{subfigure}[b]{0.15\textwidth}
         \centering
         \includegraphics[width=\textwidth]{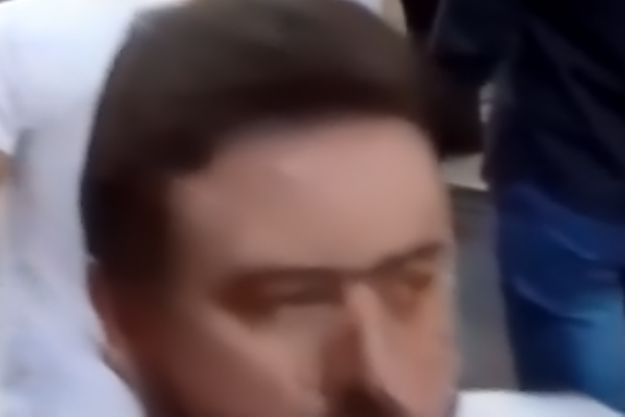}
     \end{subfigure}
     \hfill
     \begin{subfigure}[b]{0.15\textwidth}
         \centering
         \includegraphics[width=\textwidth]{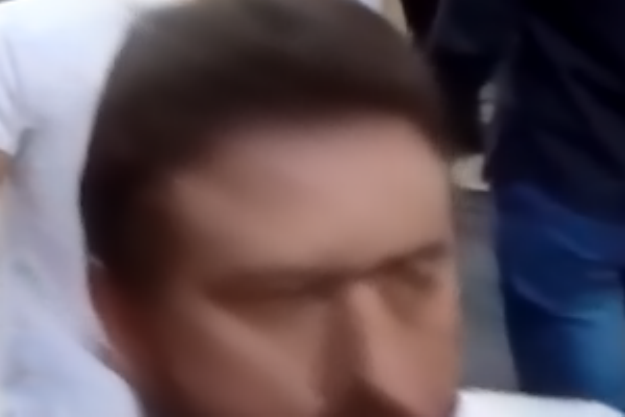}
     \end{subfigure}
     \hfill
     \begin{subfigure}[b]{0.15\textwidth}
         \centering
         \includegraphics[width=\textwidth]{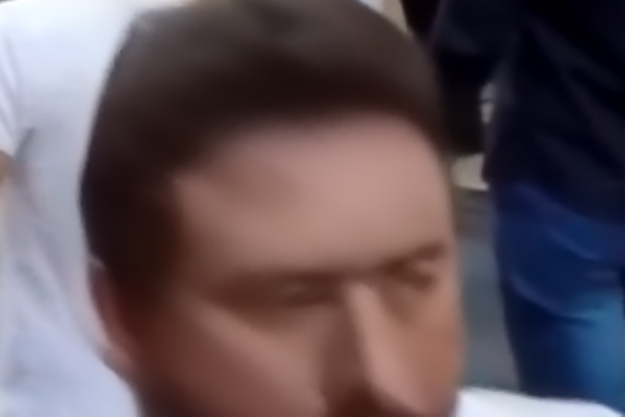}
     \end{subfigure}
     \hfill
     \begin{subfigure}[b]{0.15\textwidth}
         \centering
         \includegraphics[width=\textwidth]{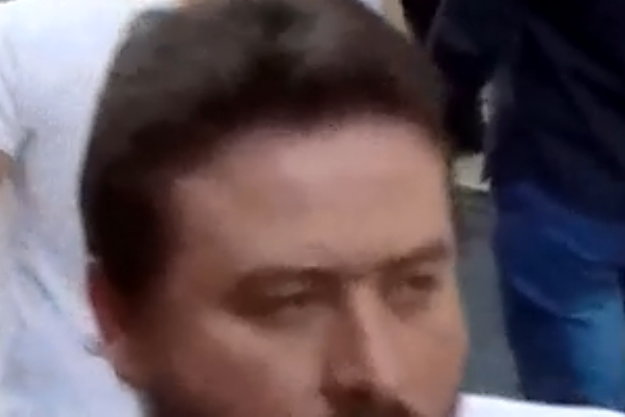}
     \end{subfigure}
     \begin{subfigure}[b]{0.18\textwidth}
         \centering
         \includegraphics[width=\textwidth]{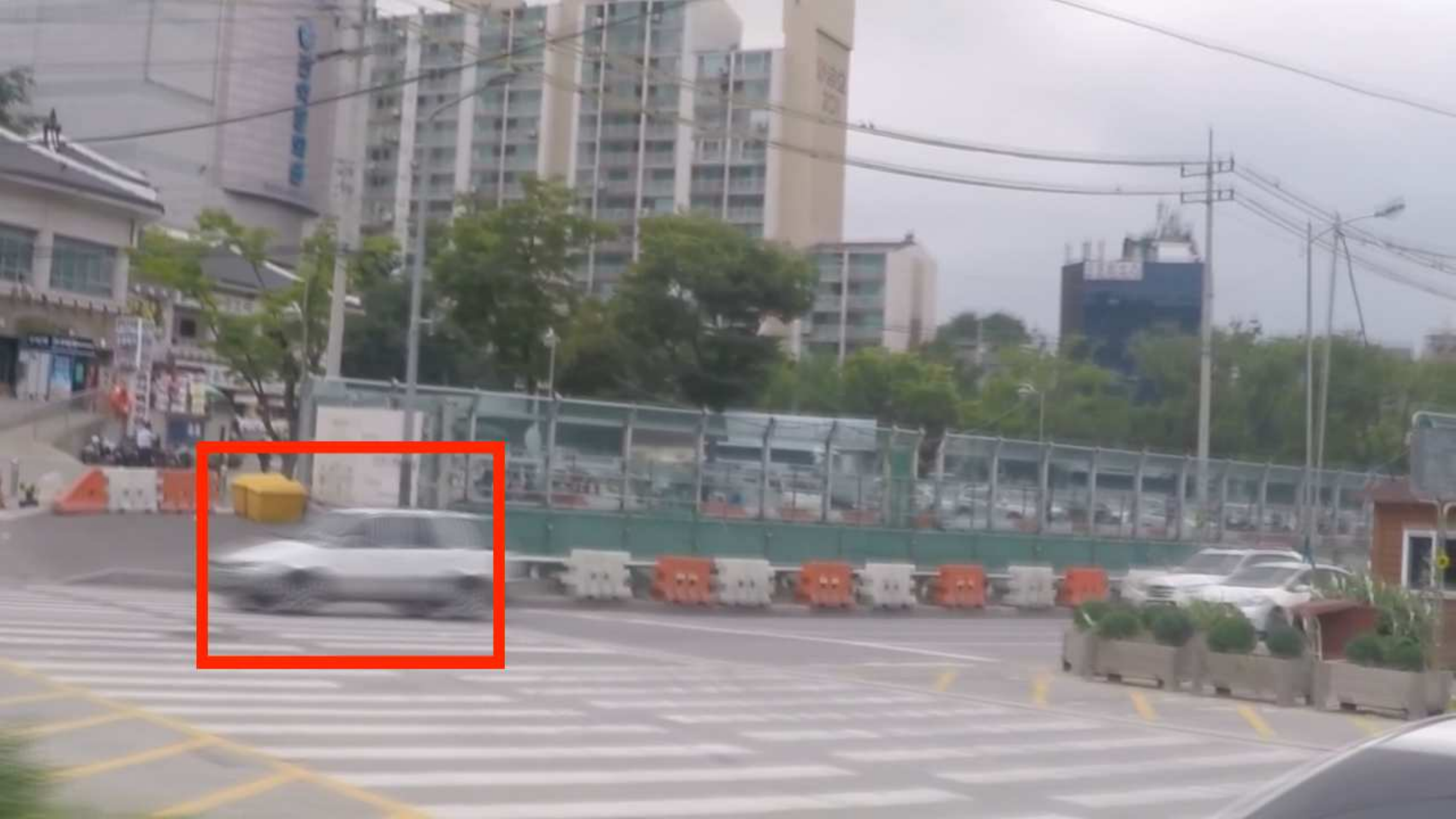}
     \end{subfigure}
     \medskip
     \hfill
     \begin{subfigure}[b]{0.15\textwidth}
         \centering
         \includegraphics[width=\textwidth]{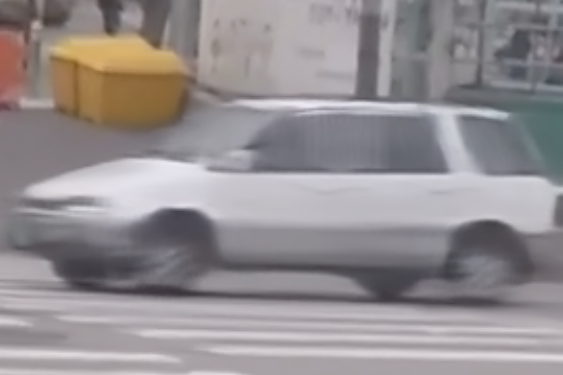}
     \end{subfigure}
     \hfill
     \begin{subfigure}[b]{0.15\textwidth}
         \centering
         \includegraphics[width=\textwidth]{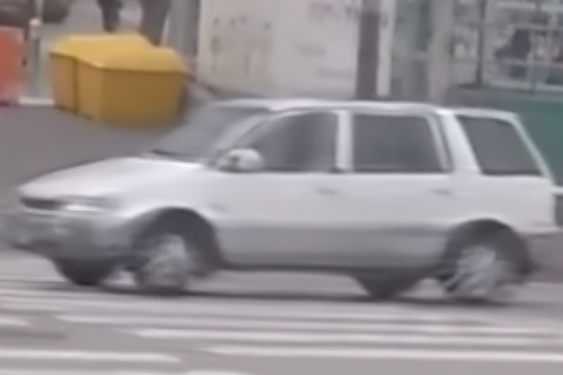}
     \end{subfigure}
     \hfill
     \begin{subfigure}[b]{0.15\textwidth}
         \centering
         \includegraphics[width=\textwidth]{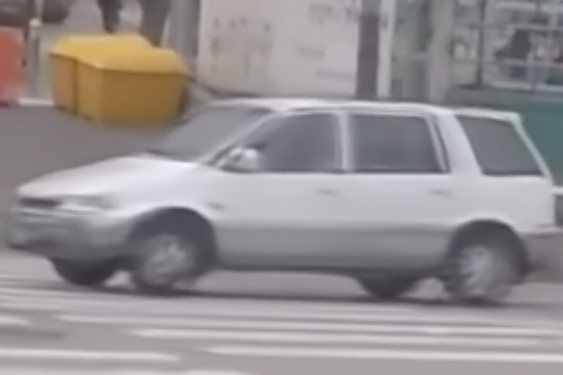}
     \end{subfigure}
     \hfill
     \begin{subfigure}[b]{0.15\textwidth}
         \centering
         \includegraphics[width=\textwidth]{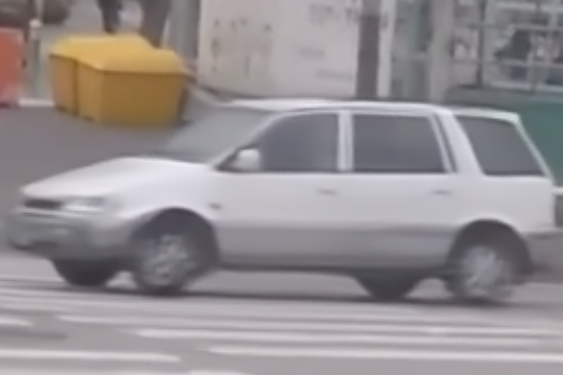}
     \end{subfigure}
     \hfill
     \begin{subfigure}[b]{0.15\textwidth}
         \centering
         \includegraphics[width=\textwidth]{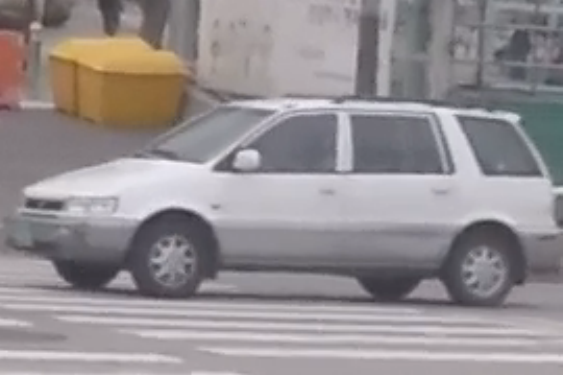}
     \end{subfigure}
     \begin{subfigure}[b]{0.18\textwidth}
         \centering
         \includegraphics[width=\textwidth]{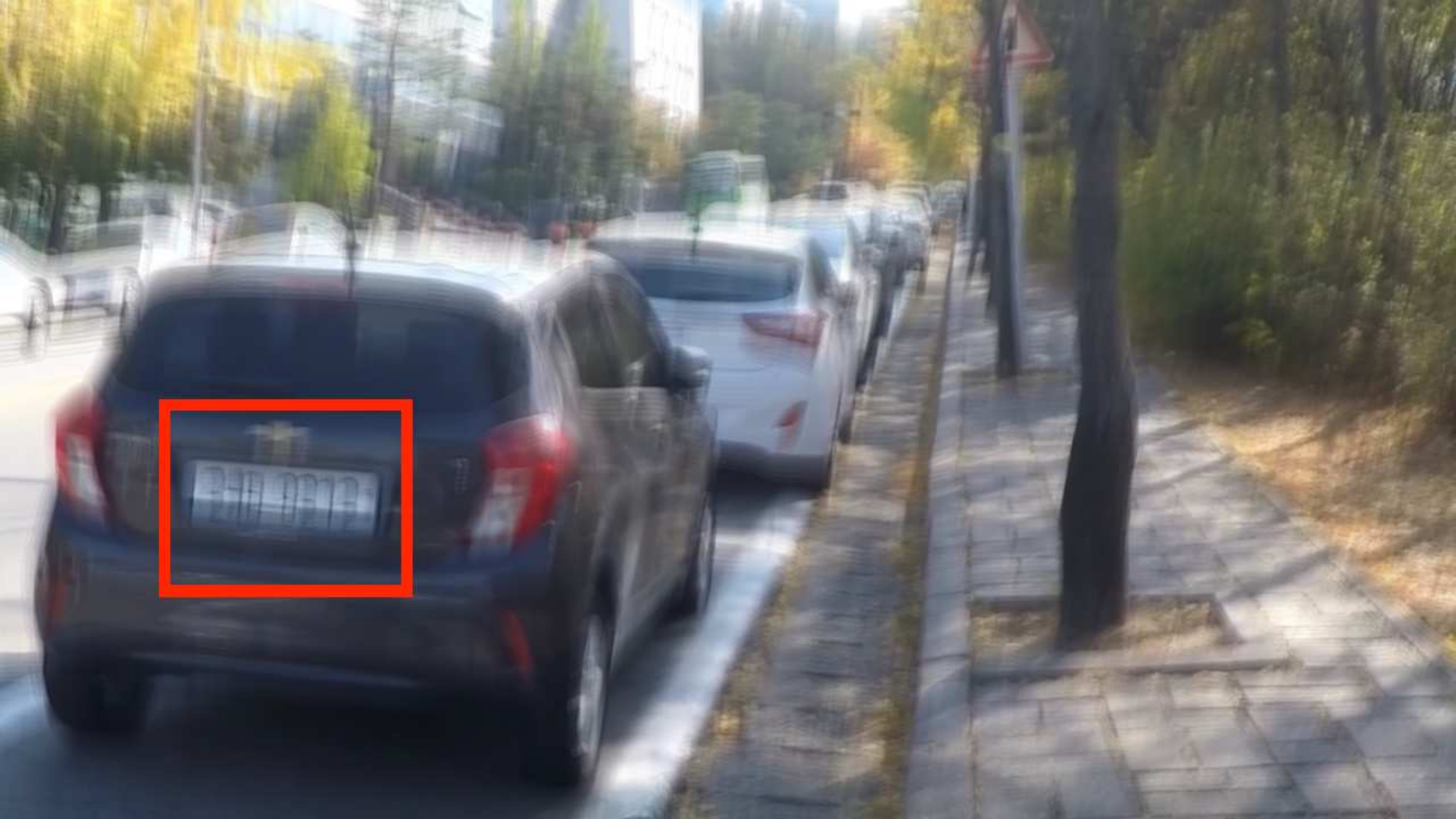}
         \caption{Input}
     \end{subfigure}
     \hfill
     \begin{subfigure}[b]{0.15\textwidth}
         \centering
         \includegraphics[width=\textwidth]{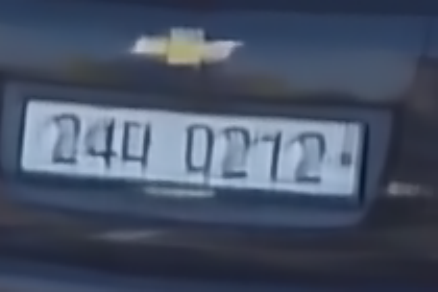}
         \caption{MPRNet}
     \end{subfigure}
     \hfill
     \begin{subfigure}[b]{0.15\textwidth}
         \centering
         \includegraphics[width=\textwidth]{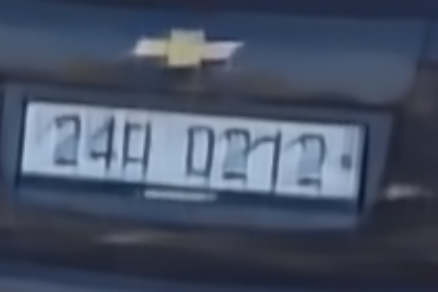}
         \caption{MIMO+}
     \end{subfigure}
     \hfill
     \begin{subfigure}[b]{0.15\textwidth}
         \centering
         \includegraphics[width=\textwidth]{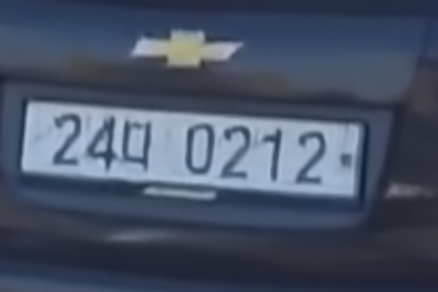}
         \caption{NAFNet}
     \end{subfigure}
     \hfill
     \begin{subfigure}[b]{0.15\textwidth}
         \centering
         \includegraphics[width=\textwidth]{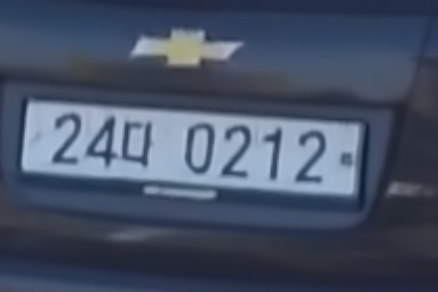}
         \caption{Ours}
     \end{subfigure}
     \hfill
     \begin{subfigure}[b]{0.15\textwidth}
         \centering
         \includegraphics[width=\textwidth]{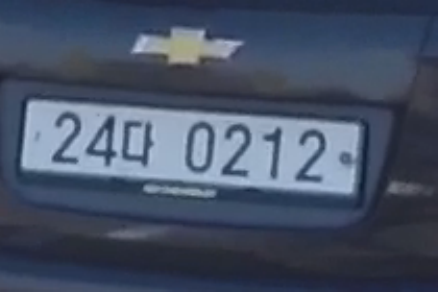}
         \caption{GT}
     \end{subfigure}
     \caption{Example results by our method (Ref-NAFNet64) and the SOTA single image deblurring methods.  }
     \label{fig:quality_comp}
\end{figure*}

Figure \ref{fig:quality_comp} shows examples of deblurred images for several challenging images that have been used in the literature. It shows the results of our Ref-NAFNet64 and the SOTA single image deblurring methods. We can see that our models achieve the best results;

\begin{figure*}[!ht]
     \centering
     \begin{subfigure}[b]{0.21\textwidth}
         \centering
         \includegraphics[width=\textwidth]{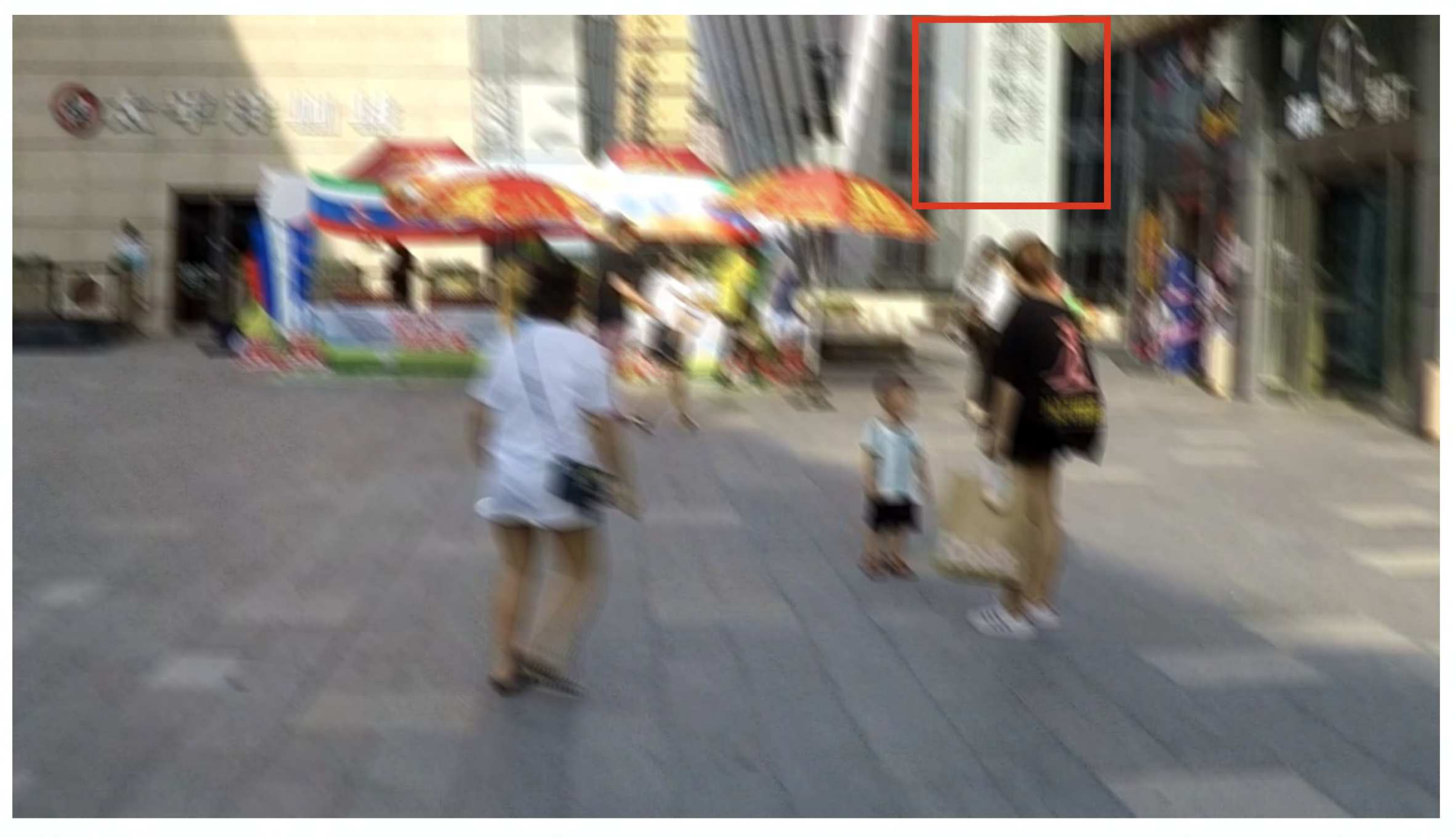}
     \end{subfigure}
     \hfill
     \begin{subfigure}[b]{0.21\textwidth}
         \centering
         \includegraphics[width=\textwidth]{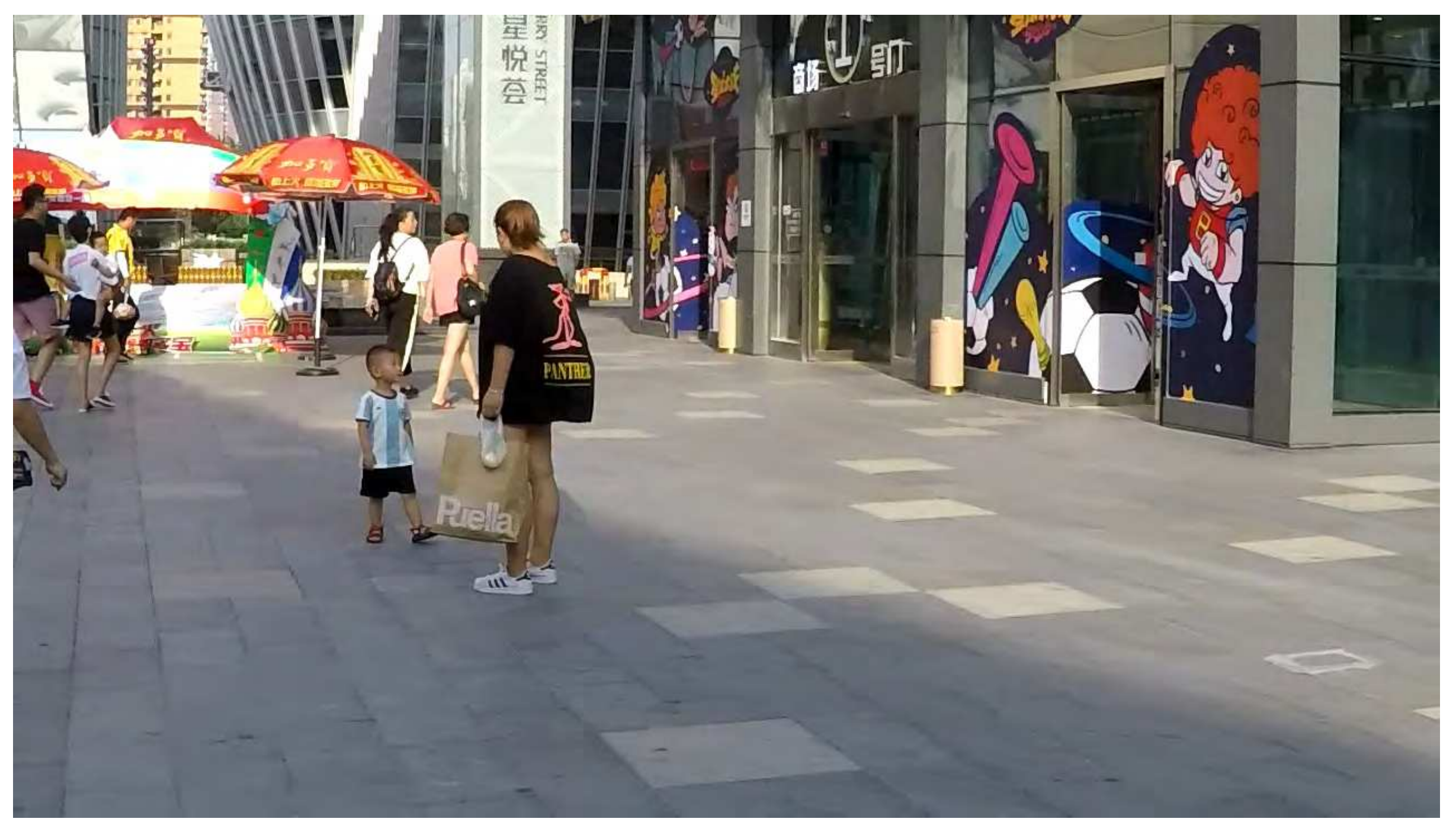}
     \end{subfigure}
     \hfill
     \begin{subfigure}[b]{0.12\textwidth}
         \centering
         \includegraphics[width=\textwidth]{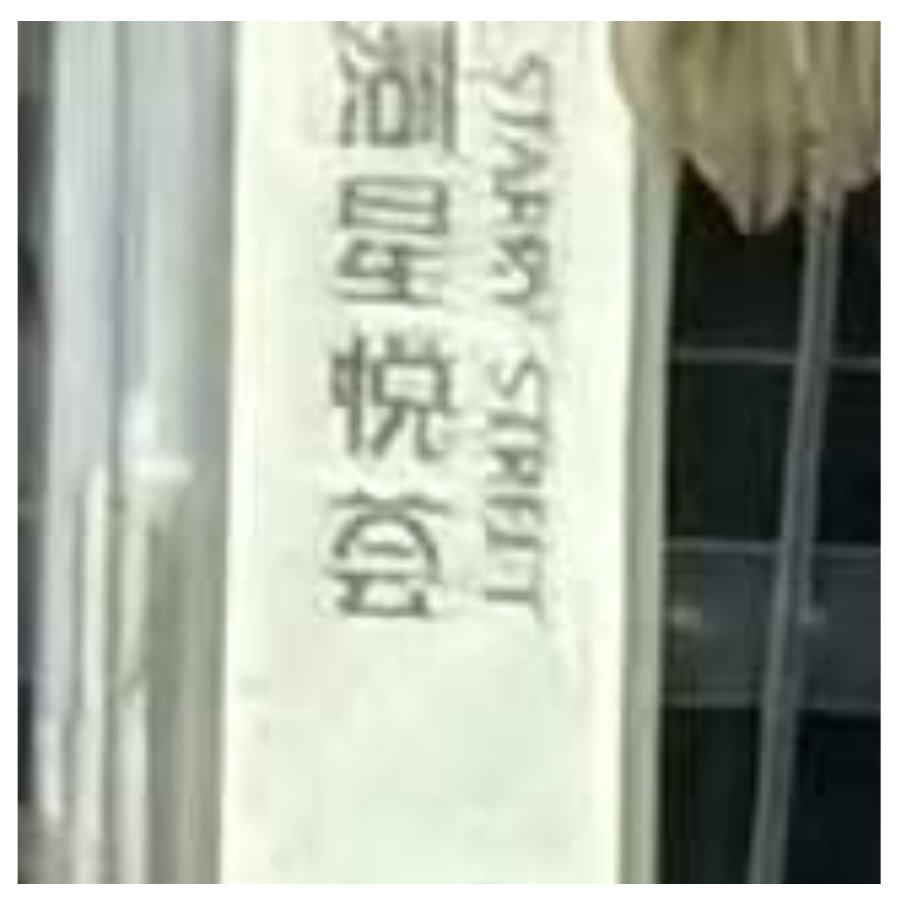}
     \end{subfigure}
     \hfill
     \begin{subfigure}[b]{0.12\textwidth}
         \centering
         \includegraphics[width=\textwidth]{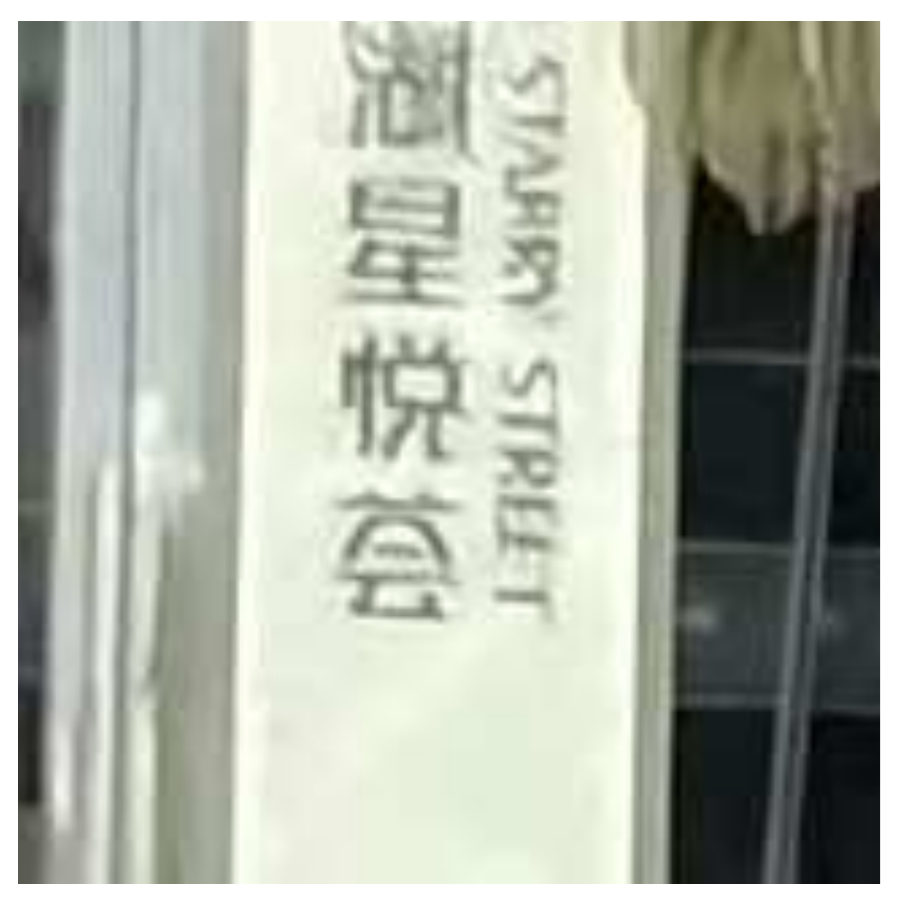}
     \end{subfigure}
     \hfill
     \begin{subfigure}[b]{0.12\textwidth}
         \centering
         \includegraphics[width=\textwidth]{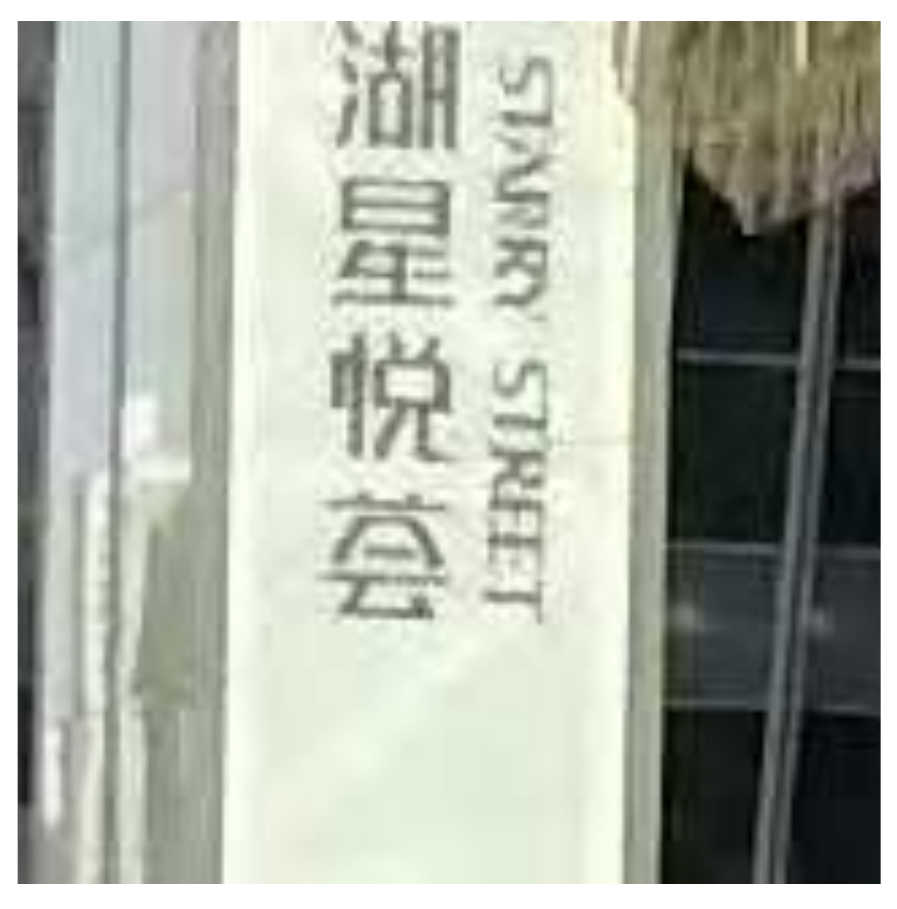}
     \end{subfigure}
     \hfill
     \begin{subfigure}[b]{0.12\textwidth}
         \centering
         \includegraphics[width=\textwidth]{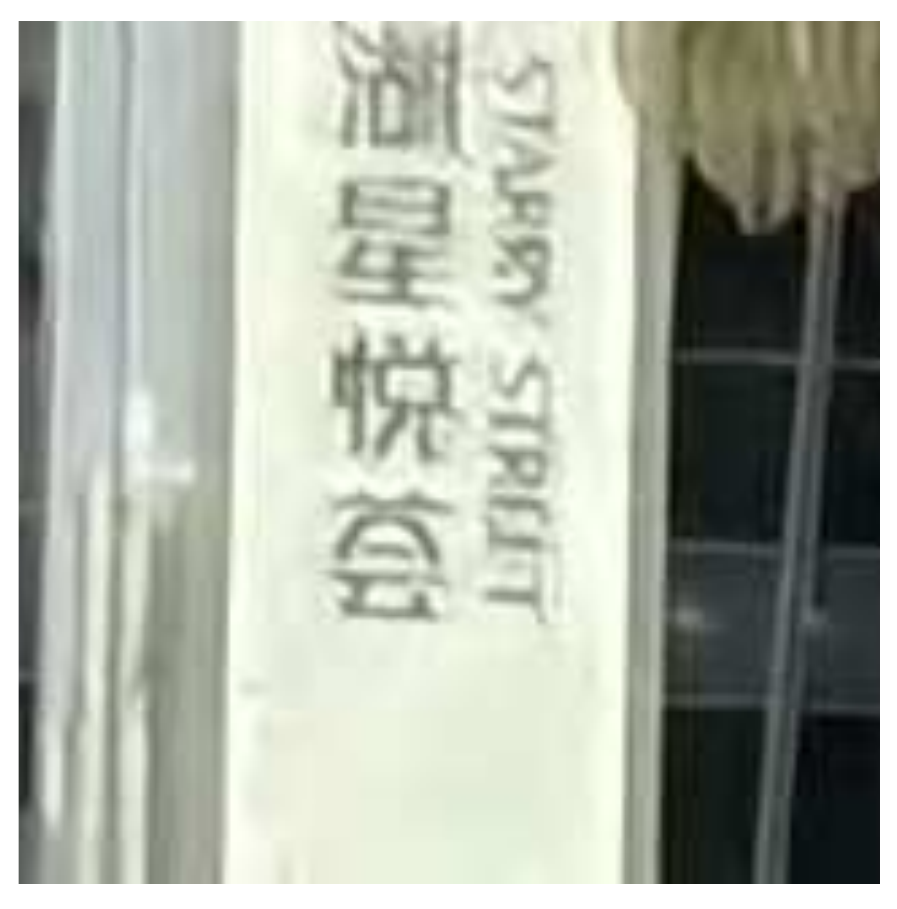}
     \end{subfigure}
     \medskip
     \begin{subfigure}[b]{0.21\textwidth}
         \centering
         \includegraphics[width=\textwidth]{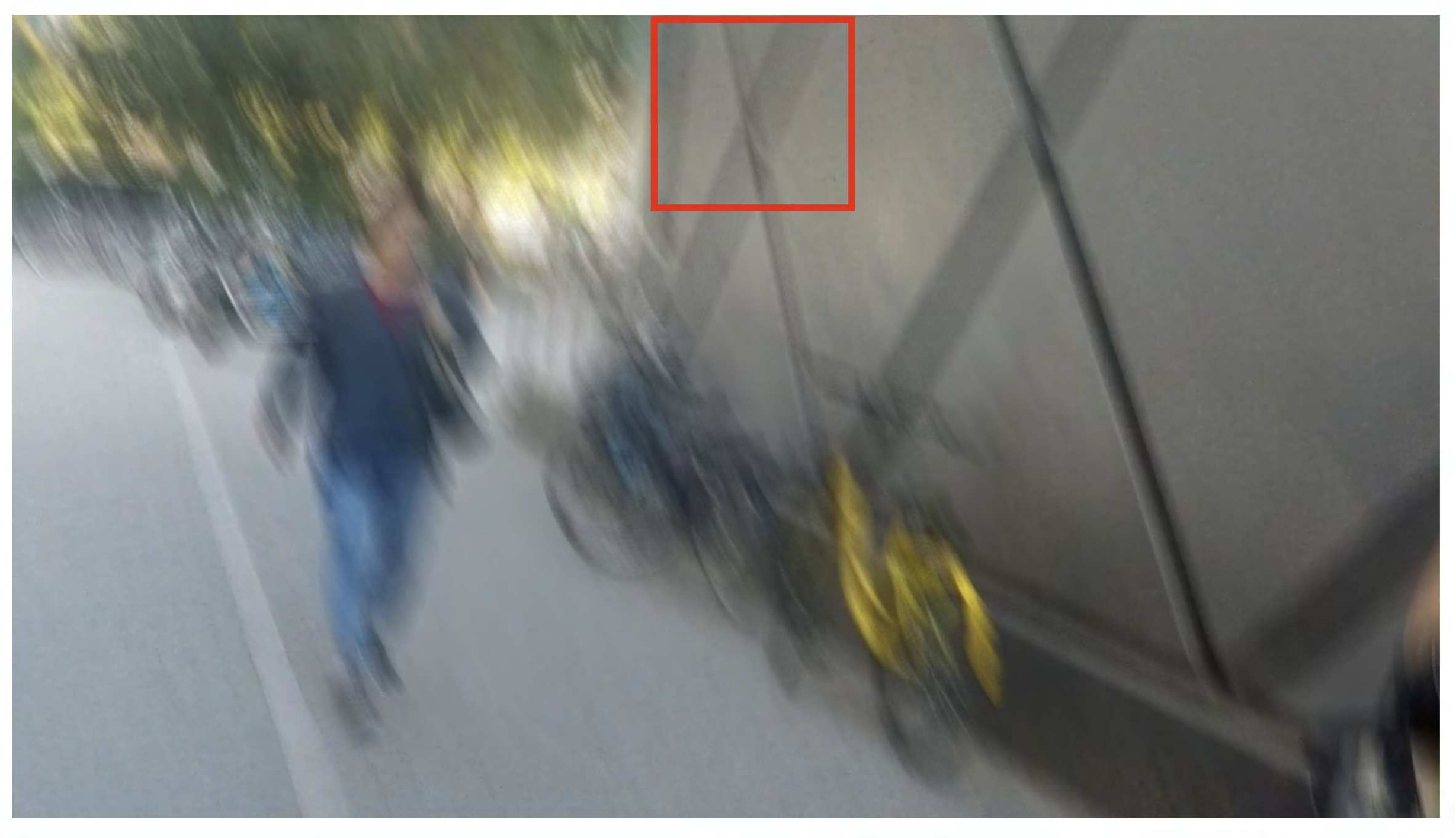}
         \caption{Blurry Input}
     \end{subfigure}
     \hfill
     \begin{subfigure}[b]{0.21\textwidth}
         \centering
         \includegraphics[width=\textwidth]{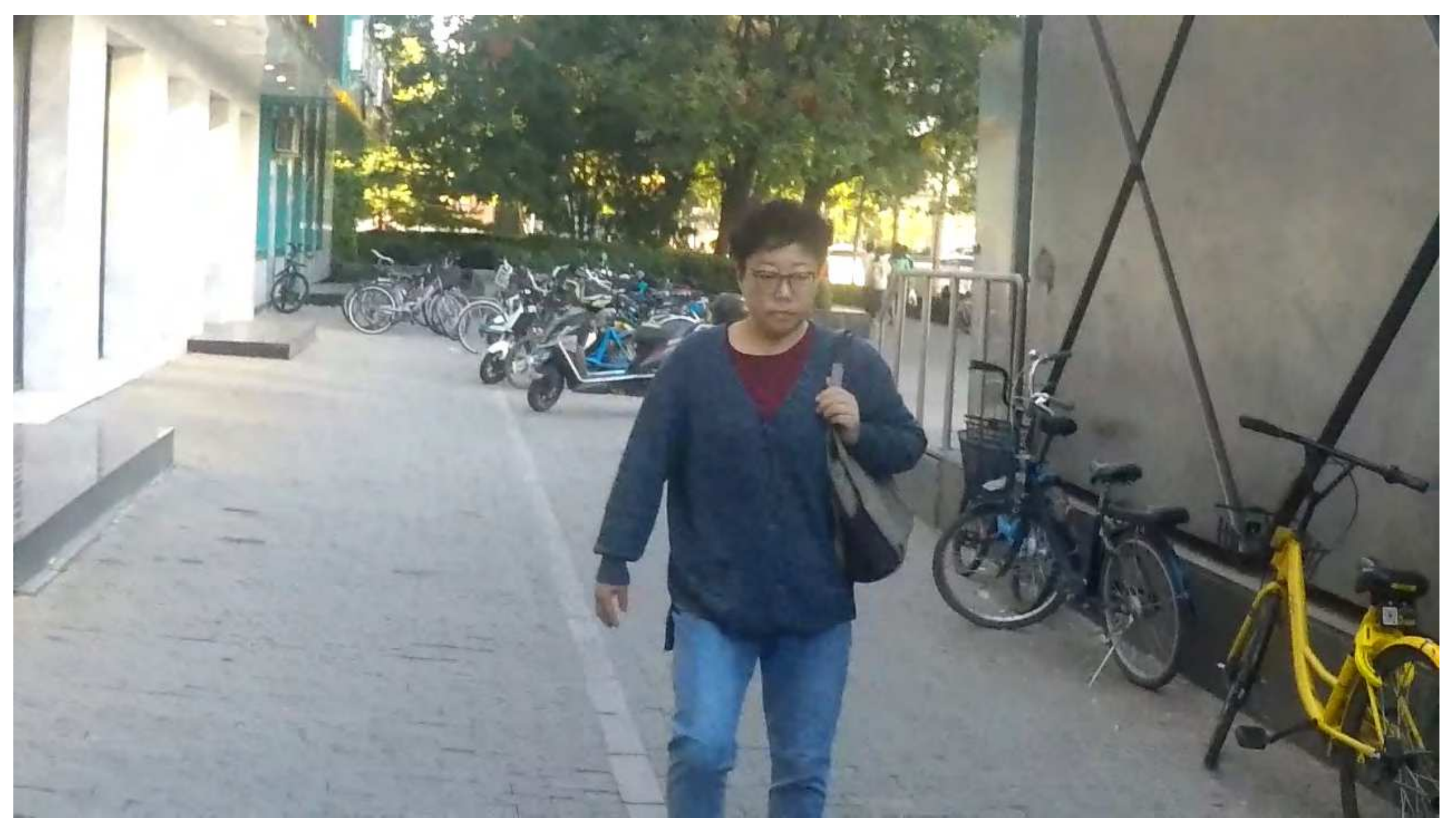}
         \caption{Reference Input}
     \end{subfigure}
     \hfill
     \begin{subfigure}[b]{0.12\textwidth}
         \centering
         \includegraphics[width=\textwidth]{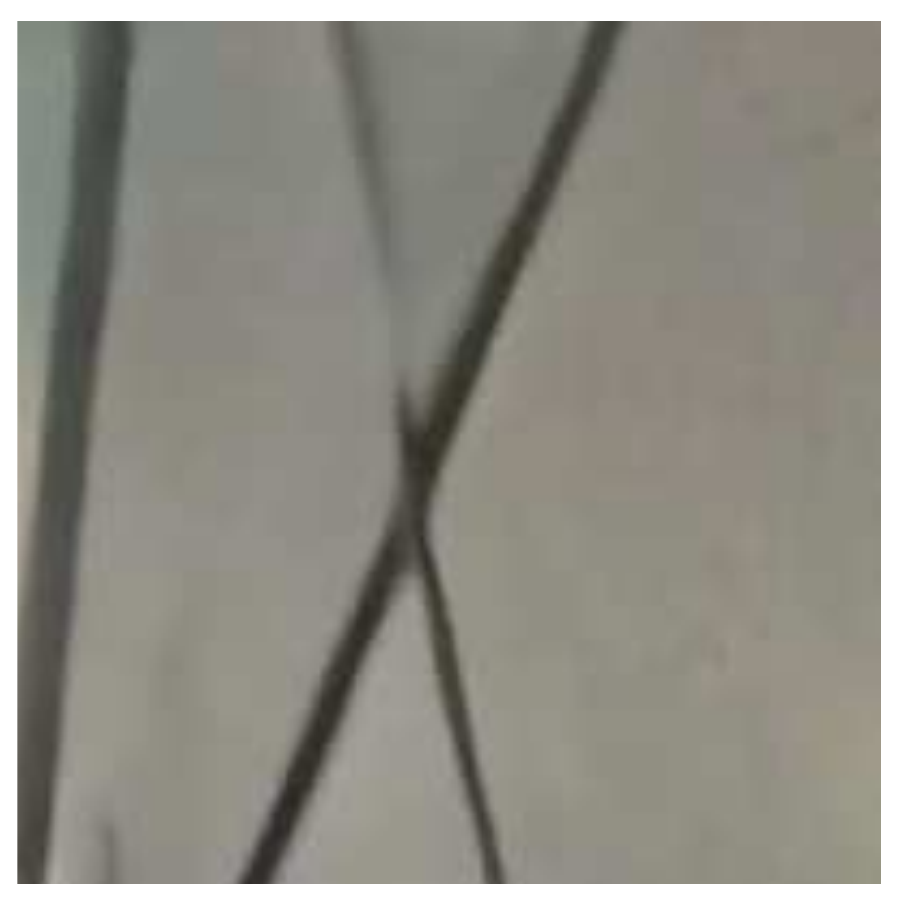}
         \caption{MIMO+}
     \end{subfigure}
     \hfill
     \begin{subfigure}[b]{0.12\textwidth}
         \centering
         \includegraphics[width=\textwidth]{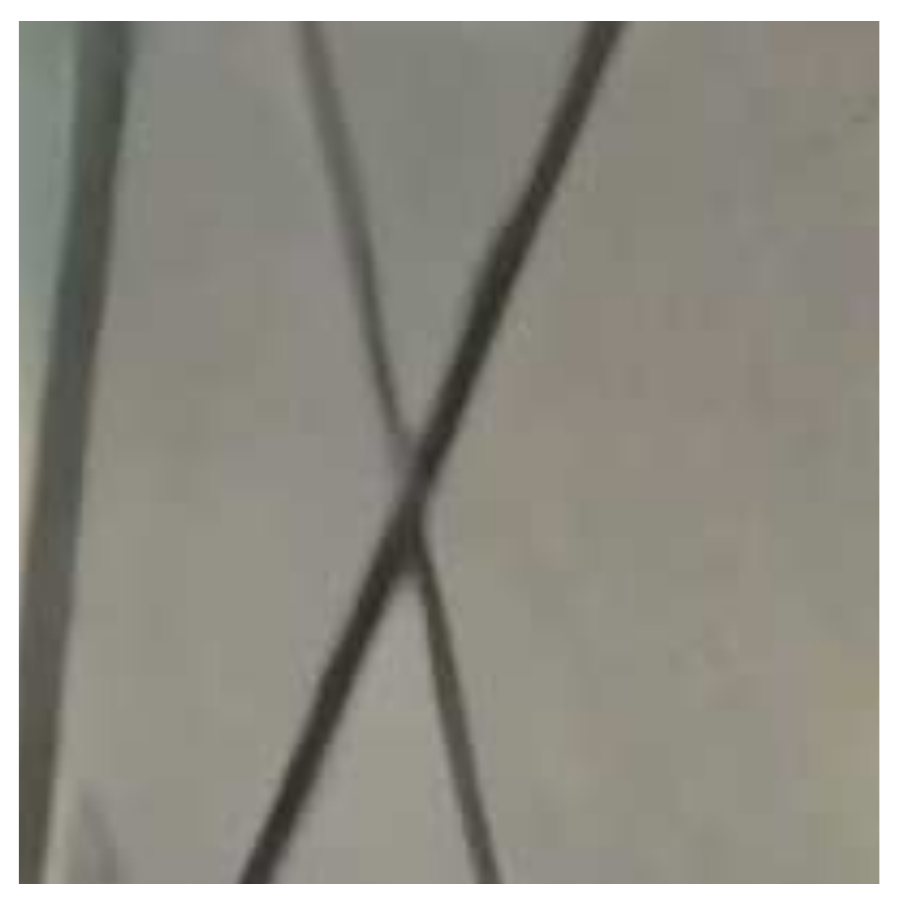}
         \caption{w/o Ref}
     \end{subfigure}
     \hfill
     \begin{subfigure}[b]{0.12\textwidth}
         \centering
         \includegraphics[width=\textwidth]{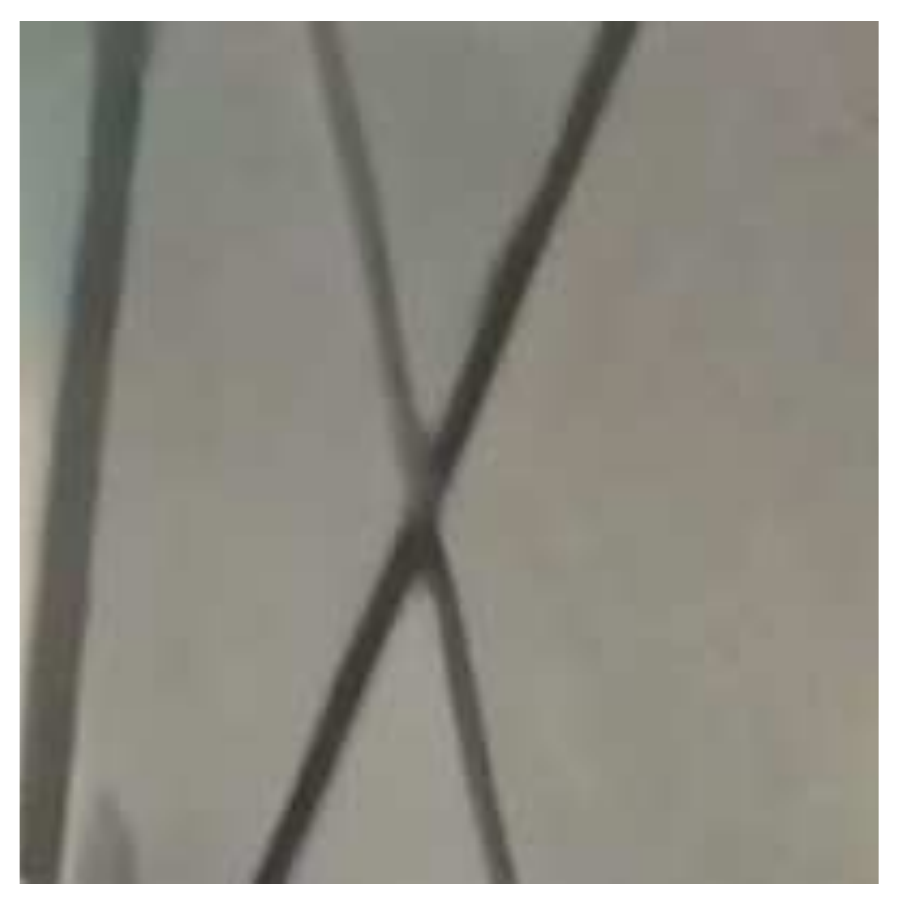}
         \caption{w Ref}
     \end{subfigure}
     \hfill
     \begin{subfigure}[b]{0.12\textwidth}
         \centering
         \includegraphics[width=\textwidth]{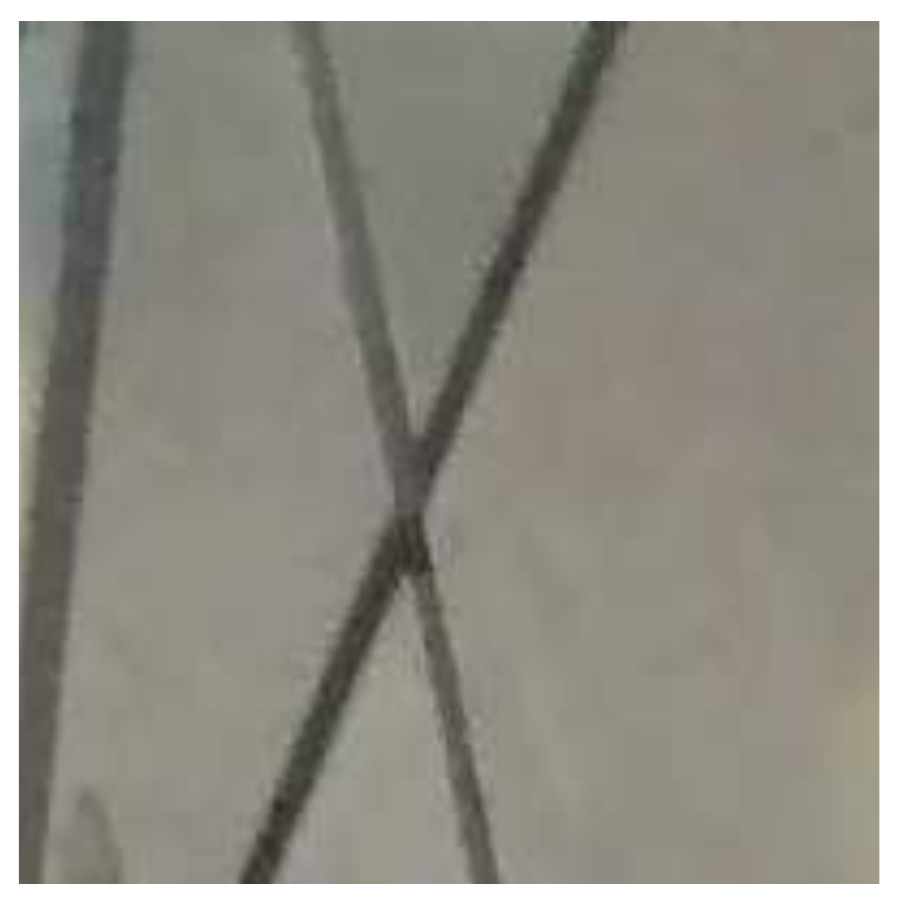}
         \caption{GT}
     \end{subfigure}
     \caption{Comparison between original MIMO-UNet+~\cite{cho2021rethinking} and our Ref-MIMO-UNet+ on an image from HIDE~\cite{shen2019human} }
     \label{fig:comp_ref1}
\end{figure*}

\begin{figure}[!ht]
\begin{minipage}{0.98\columnwidth}
     \centering
     \begin{subfigure}[b]{0.47\textwidth}
         \centering
         \includegraphics[width=\textwidth]{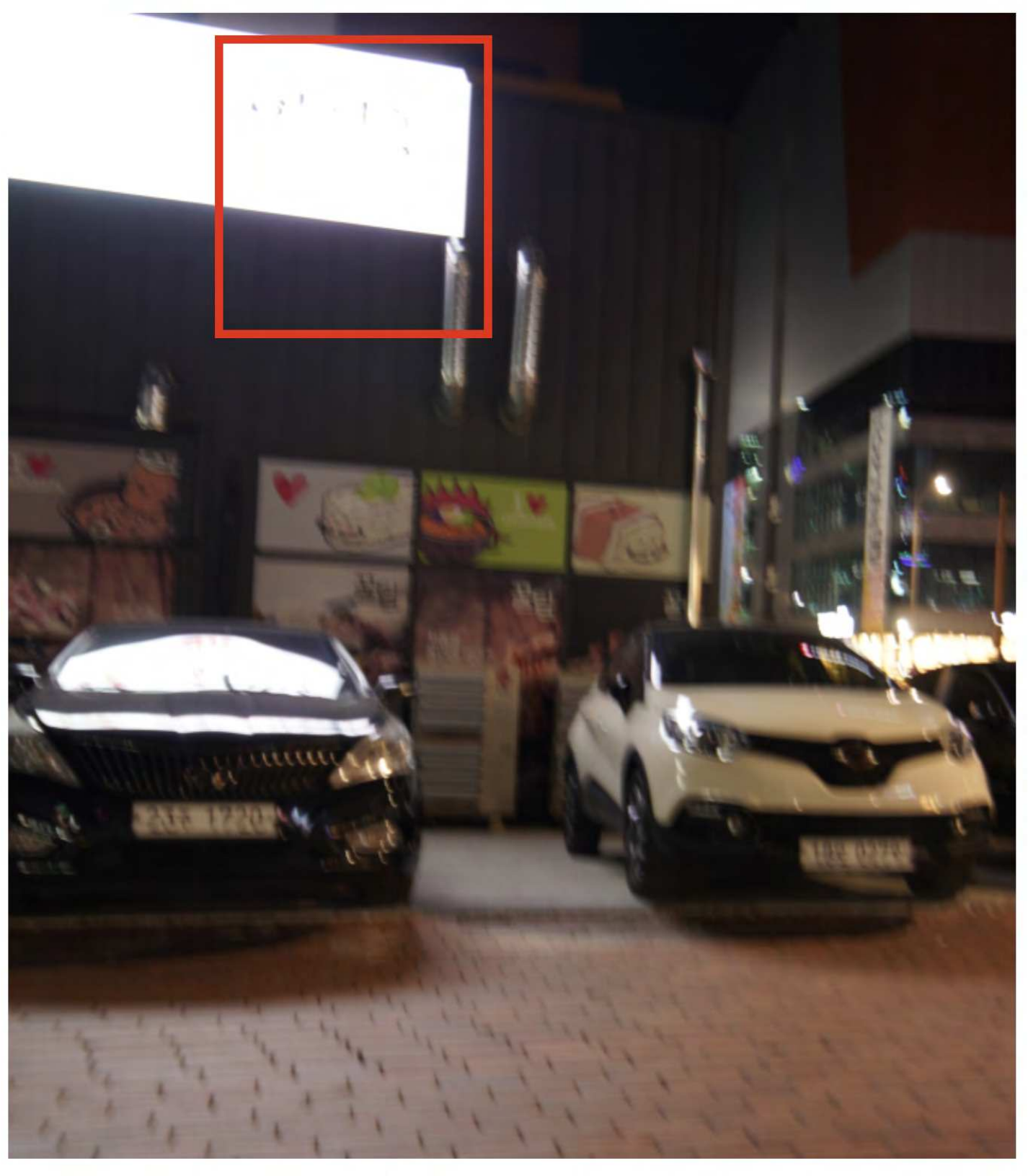}
         \caption{Blurry Input}
     \end{subfigure}
     \begin{subfigure}[b]{0.47\textwidth}
         \centering
         \includegraphics[width=\textwidth]{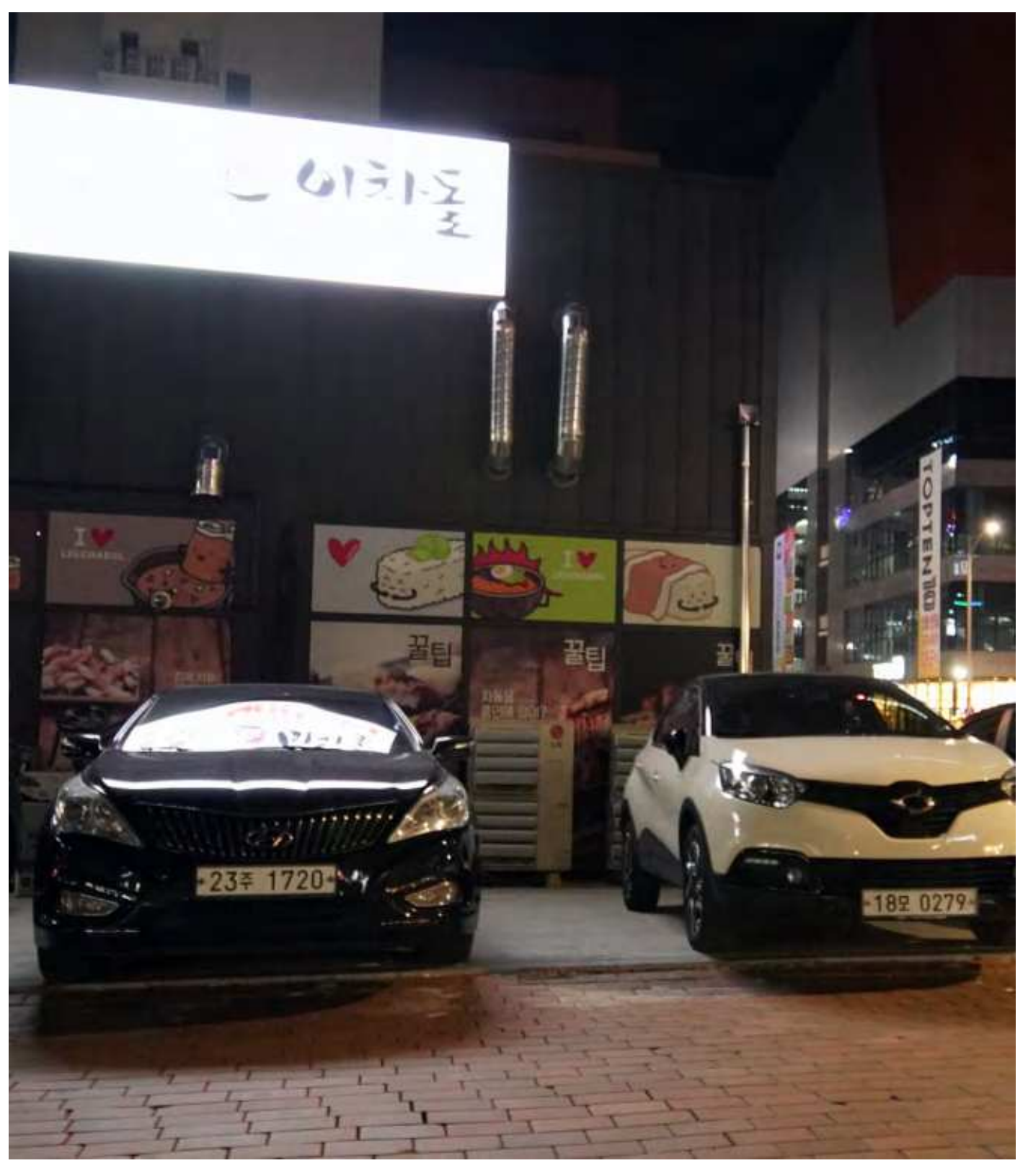}
         \caption{Reference Input}
     \end{subfigure}
     \hfill
     \medskip
     \begin{subfigure}[b]{0.24\textwidth}
         \centering
         \includegraphics[width=\textwidth]{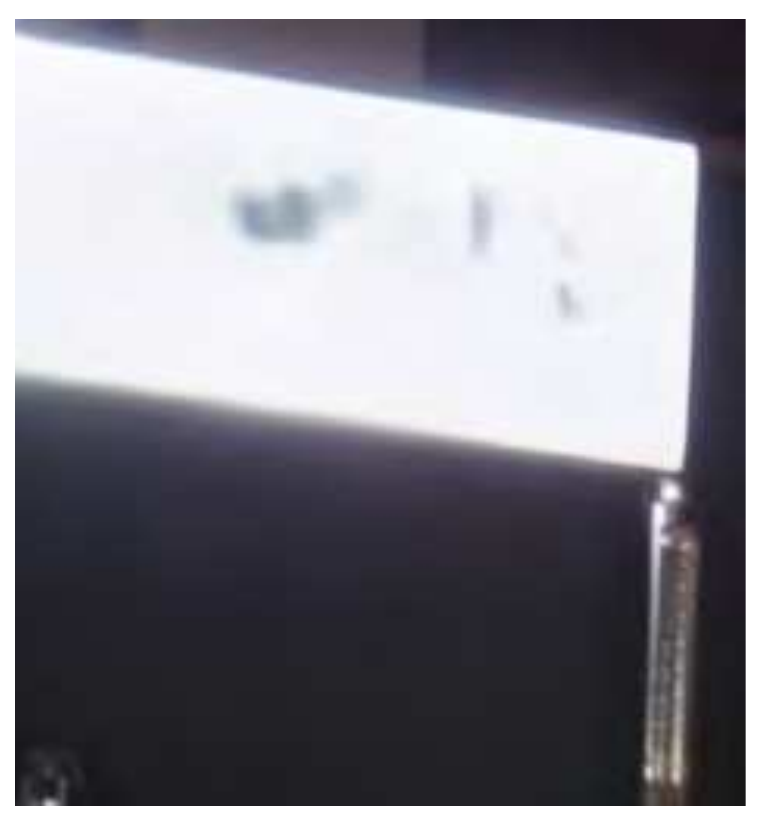}
         \caption{MIMO+}
     \end{subfigure}
     \hfill
     \begin{subfigure}[b]{0.24\textwidth}
         \centering
         \includegraphics[width=\textwidth]{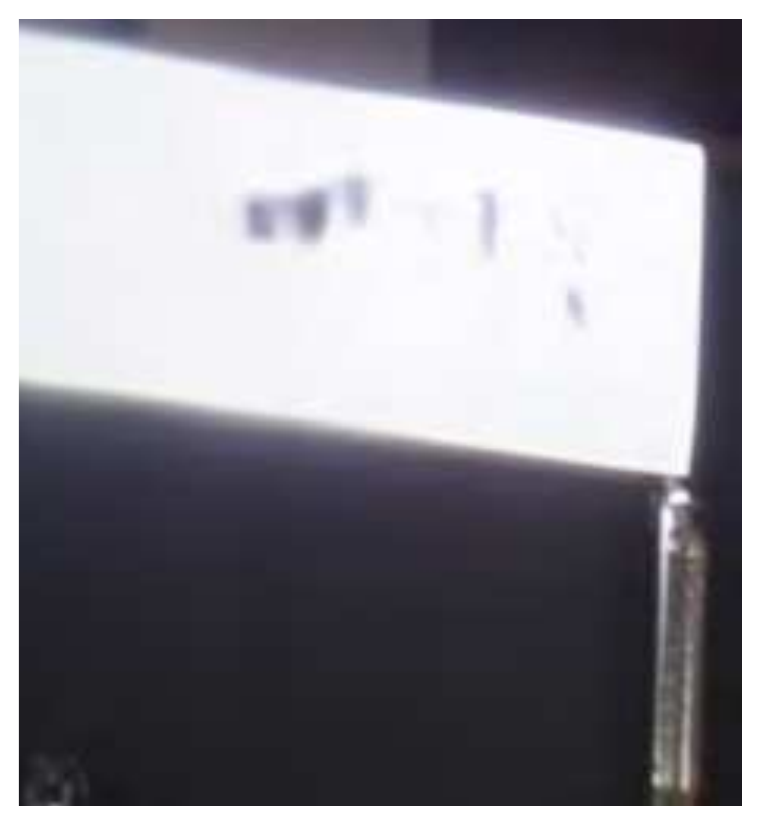}
         \caption{w/o Ref}
     \end{subfigure}
     \hfill
     \begin{subfigure}[b]{0.24\textwidth}
         \centering
         \includegraphics[width=\textwidth]{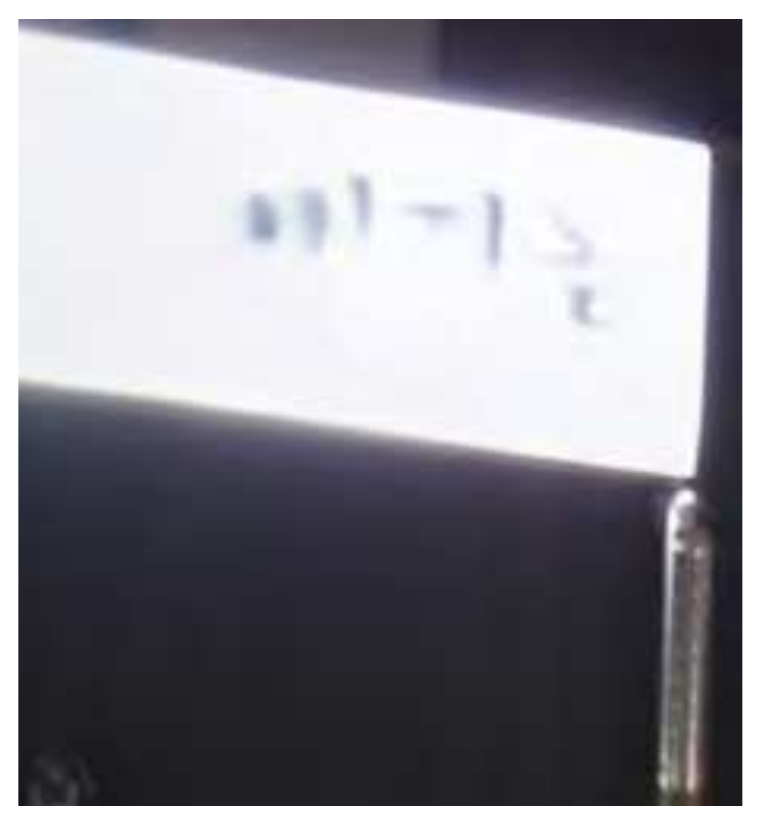}
         \caption{w Ref}
     \end{subfigure}
     \hfill
     \begin{subfigure}[b]{0.24\textwidth}
         \centering
         \includegraphics[width=\textwidth]{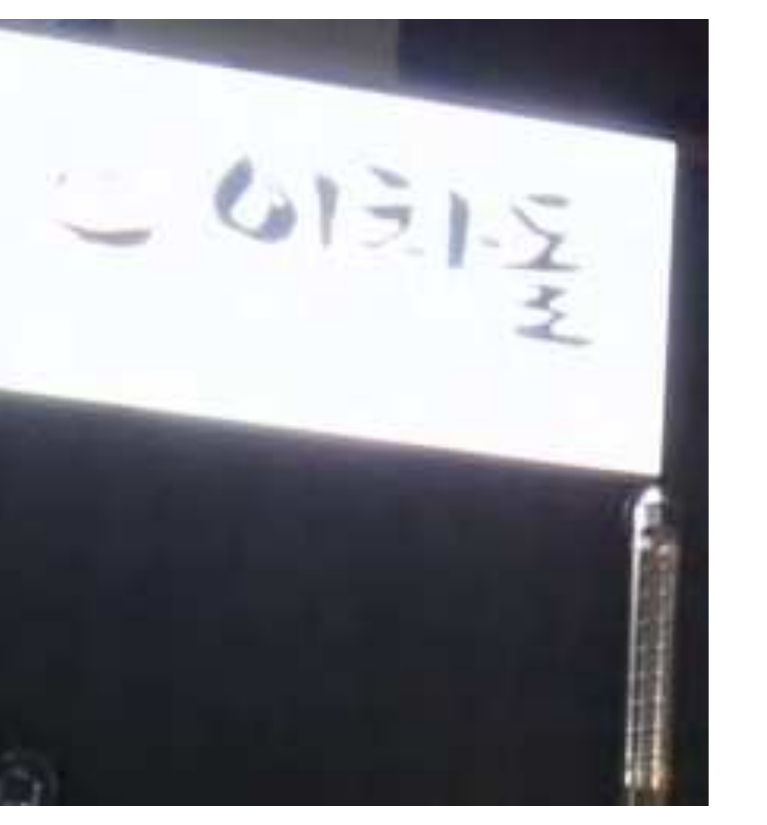}
         \caption{GT}
     \end{subfigure}
     \caption{Comparison between original MIMO-UNet+~\cite{cho2021rethinking} and our Ref-MIMO-UNet+ on an image from RealBlur~\cite{rim2020real} }
     \label{fig:comp_ref3}
     \end{minipage}
\end{figure}

\begin{figure}[!ht]
\begin{minipage}{0.98\columnwidth}

     \centering
     \begin{subfigure}[b]{0.47\textwidth}
         \centering
         \includegraphics[width=\textwidth]{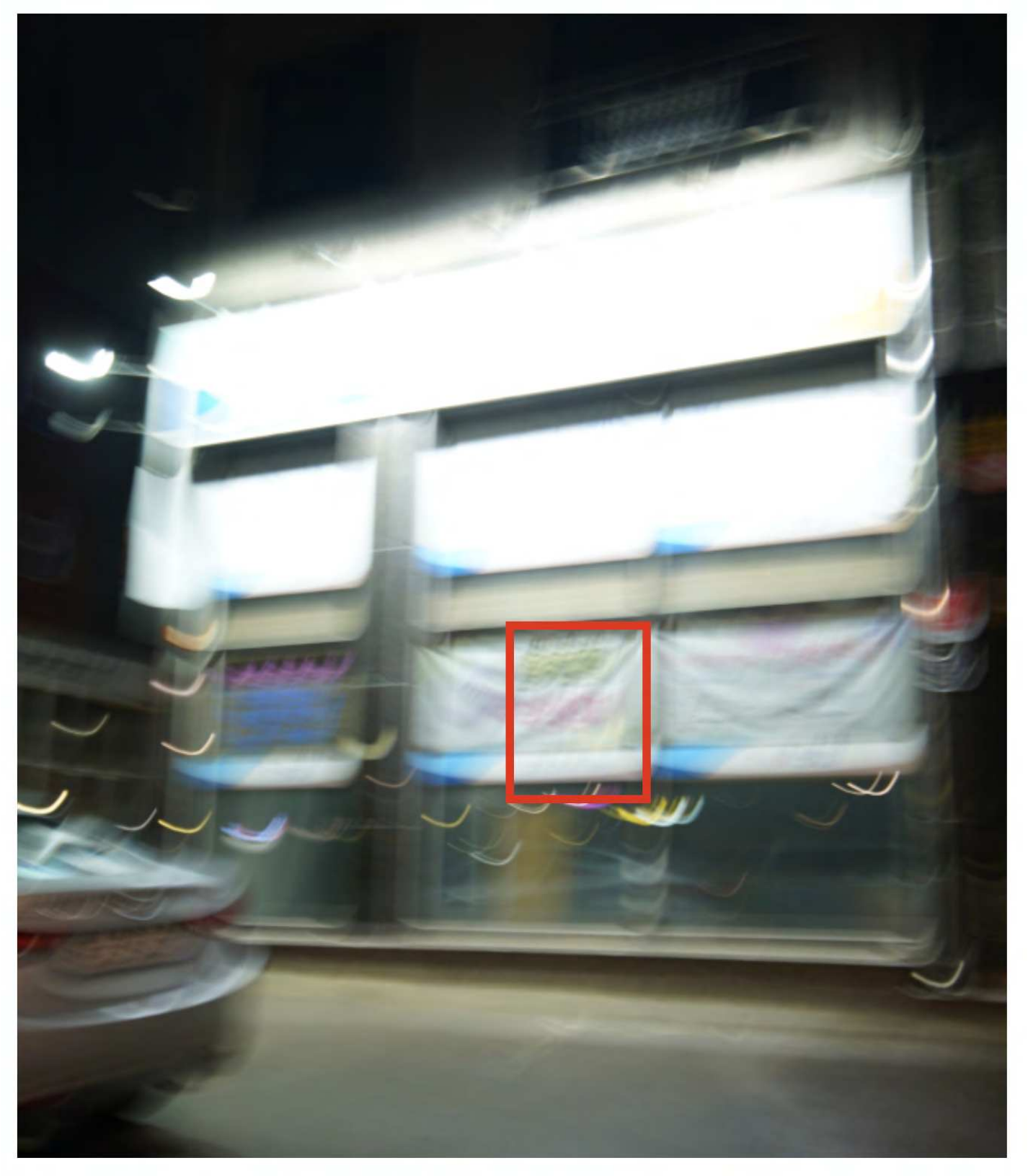}
         \caption{Blurry Input}
     \end{subfigure}
     \begin{subfigure}[b]{0.47\textwidth}
         \centering
         \includegraphics[width=\textwidth]{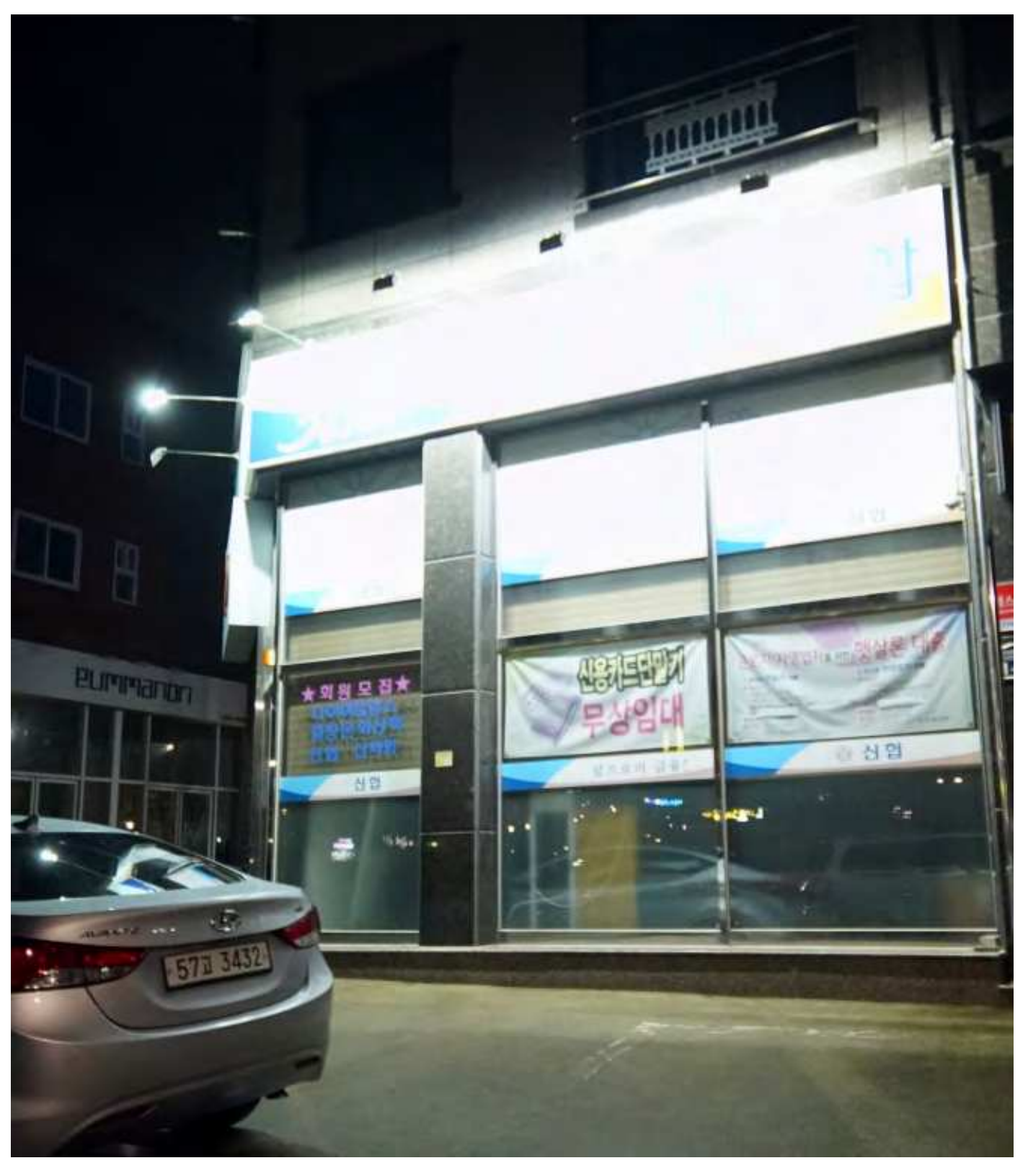}
         \caption{Reference Input}
     \end{subfigure}
     \hfill
     \medskip
     \begin{subfigure}[b]{0.24\textwidth}
         \centering
         \includegraphics[width=\textwidth]{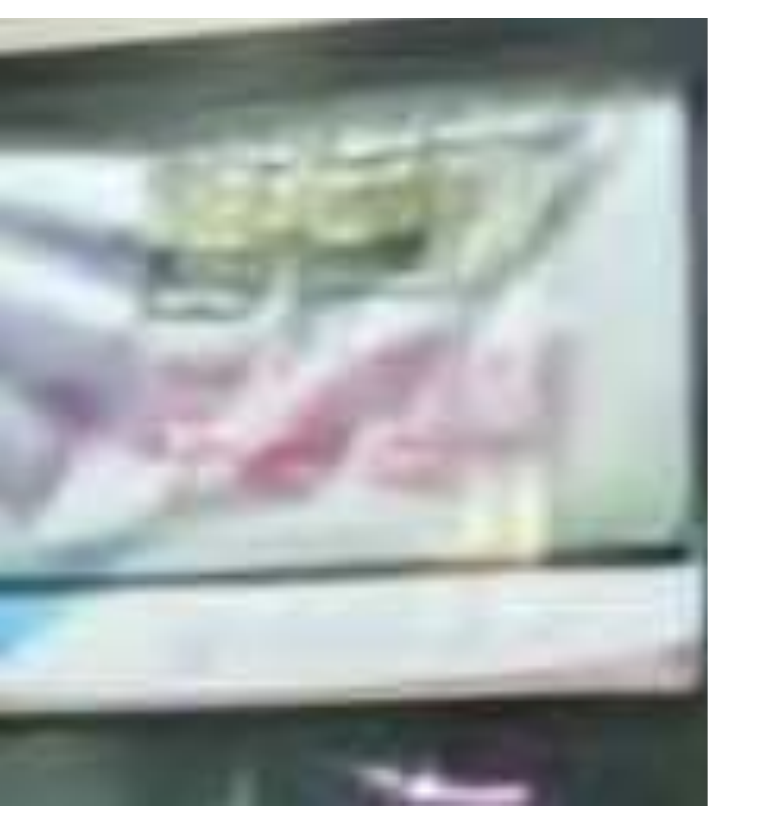}
         \caption{MIMO+}
     \end{subfigure}
     \hfill
     \begin{subfigure}[b]{0.24\textwidth}
         \centering
         \includegraphics[width=\textwidth]{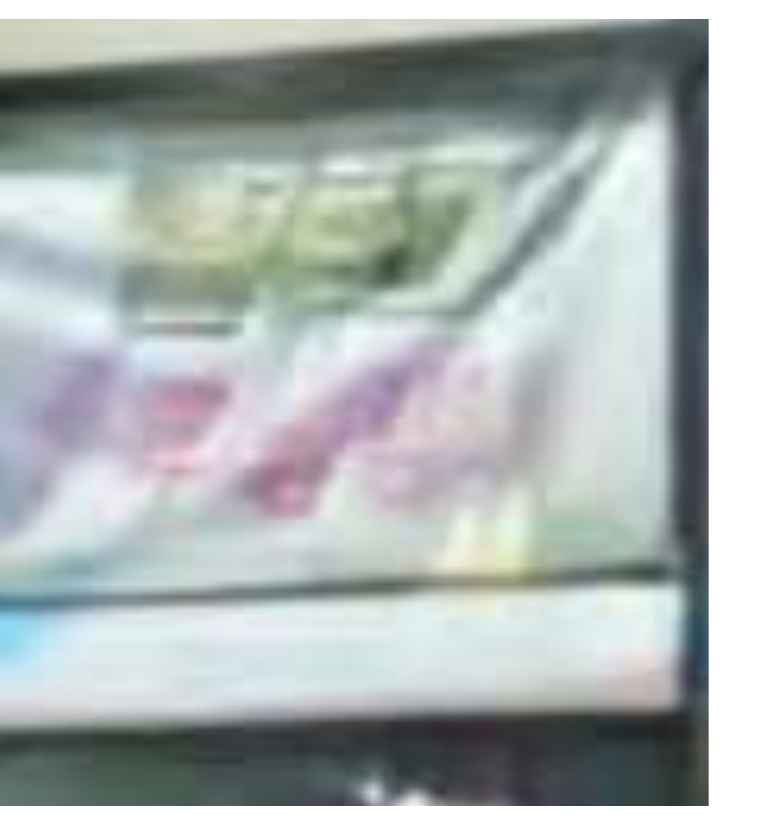}
         \caption{w/o Ref}
     \end{subfigure}
     \hfill
     \begin{subfigure}[b]{0.24\textwidth}
         \centering
         \includegraphics[width=\textwidth]{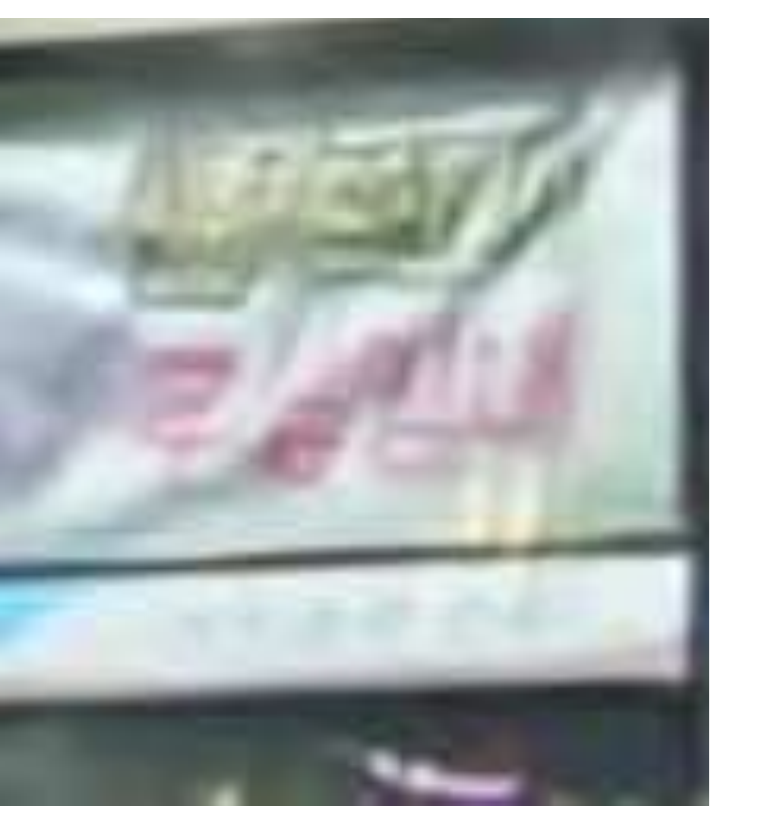}
         \caption{w Ref}
     \end{subfigure}
     \hfill
     \begin{subfigure}[b]{0.24\textwidth}
         \centering
         \includegraphics[width=\textwidth]{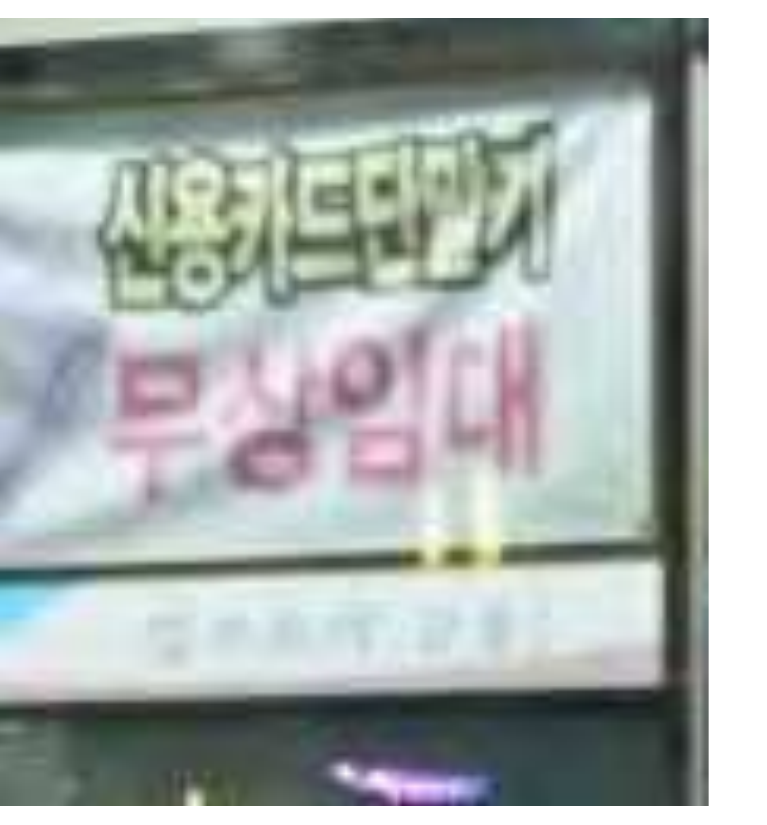}
         \caption{GT}
     \end{subfigure}
     \caption{Comparison between original MIMO-UNet+~\cite{cho2021rethinking} and our Ref-MIMO-UNet+ on an image from RealBlur~\cite{rim2020real} }
     \label{fig:comp_ref4}
     \end{minipage}
\end{figure}

Figures~\ref{fig:comp_ref1}, ~\ref{fig:comp_ref3}, and ~\ref{fig:comp_ref4} compare the original MIMO-UNet+\cite{cho2021rethinking} and our modified Ref-MIMO-UNet+ on the HIDE~\cite{shen2019human} and RealBlur~\cite{rim2020real} datasets. We also compared the deblurred results of our model without reference, which shows the effectiveness of using a reference image. 

\subsection{Other Results} 
\subsubsection{Ablation Study of Feature Matching }
We employ a multi-output architecture and create a feature matching module that conducts matching on the intermediate output. In order to assess the efficacy of this approach, we compare it to models that directly perform feature matching on the blurry input and reference input. The results in Table~\ref{table:comp_inter} indicate that utilizing the intermediate output for feature matching yields a 0.1dB improvement in PSNR performance on GOPRO dataset. This demonstrates the effectiveness of the proposed design.

\setlength{\tabcolsep}{6pt}
\begin{table}[!ht]
\begin{minipage}{0.98\columnwidth}
\caption{Ablatino study of feature matching on GOPRO }
\label{table:comp_inter}
\begin{center}
\begin{tabular}{lccc}
\hline
Methods & with inter  &  PSNR & SSIM  \\\cline{2-4}
\hline
Ref-DeblurNet           &  \XSolidBrush     & 31.56 & 0.950 \\
Ref-DeblurNet           &  \CheckmarkBold   & 31.68& 0.951 \\
Ref-MIMO-UNet           &  \XSolidBrush     & 32.42 & 0.957 \\
Ref-MIMO-UNet           & \CheckmarkBold    & 32.53 & 0.958 \\
Ref-NAFNet32            &   \XSolidBrush      & 32.12 & 0.963\\
Ref-NAFNet32            &  \CheckmarkBold   &  33.22 & 0.964\\
\hline
\end{tabular}
\end{center}
\end{minipage}
\end{table}

\subsubsection{Detailed Comparison and Results on Different SOTA methods}
 To further analyze the effectiveness of proposed methods on other SOTA deblurring models, we conduct several experiments with different backbones and test different configurations of the feature fusion module. We report their results in Table \ref{table:effectiveness};
fusion num. $=1, 2$ means that the reference features are fused with those of the blurry image on a scale of 0.5, 0.25. We can see that fusing on more scales achieves a better result. 
The effectiveness of the proposed coarse-to-fine design is showcased. 
 
\setlength{\tabcolsep}{3pt}
\begin{table}[!ht]
\begin{minipage}{0.98\columnwidth}
\caption{More results on the GOPRO dataset. }
\label{table:effectiveness}
\begin{center}
\begin{tabular}{lcccc}
\hline
Methods & Fusion num.  & Ref & PSNR & SSIM  \\\cline{2-5}
\hline
DeblurNet~\cite{zhou2019davanet}             & - & \XSolidBrush      & 31.22 & 0.946 \\
Ref-DeblurNet              & 1 & \CheckmarkBold      & 31.45 & 0.958 \\
Ref-DeblurNet         & 1,2 &  \CheckmarkBold   & 31.68& 0.951 \\
MIMO-UNet\cite{cho2021rethinking} & - & \XSolidBrush & 32.10 & 0.954 \\
Ref-MIMO-UNet   & 1 & \CheckmarkBold    & 32.33 & 0.957 \\
Ref-MIMO-UNet   & 1,2 & \CheckmarkBold    & 32.53 & 0.958 \\
NAFNet32\cite{chen2022simple}          & - & \XSolidBrush      & 32.90 & 0.960 \\
Ref-NAFNet32     & 1 & \CheckmarkBold      & 32.07 & 0.962\\
Ref-NAFNet32      & 1,2 & \CheckmarkBold  &  33.22 & 0.964\\
\hline
\end{tabular}
\end{center}
\end{minipage}
\end{table}

\subsubsection{Impacts of Reference Selection}
\begin{figure*}[!ht]
    \centering
     \begin{subfigure}[b]{0.22\textwidth}
         \centering
         \includegraphics[height=0.3\textheight]{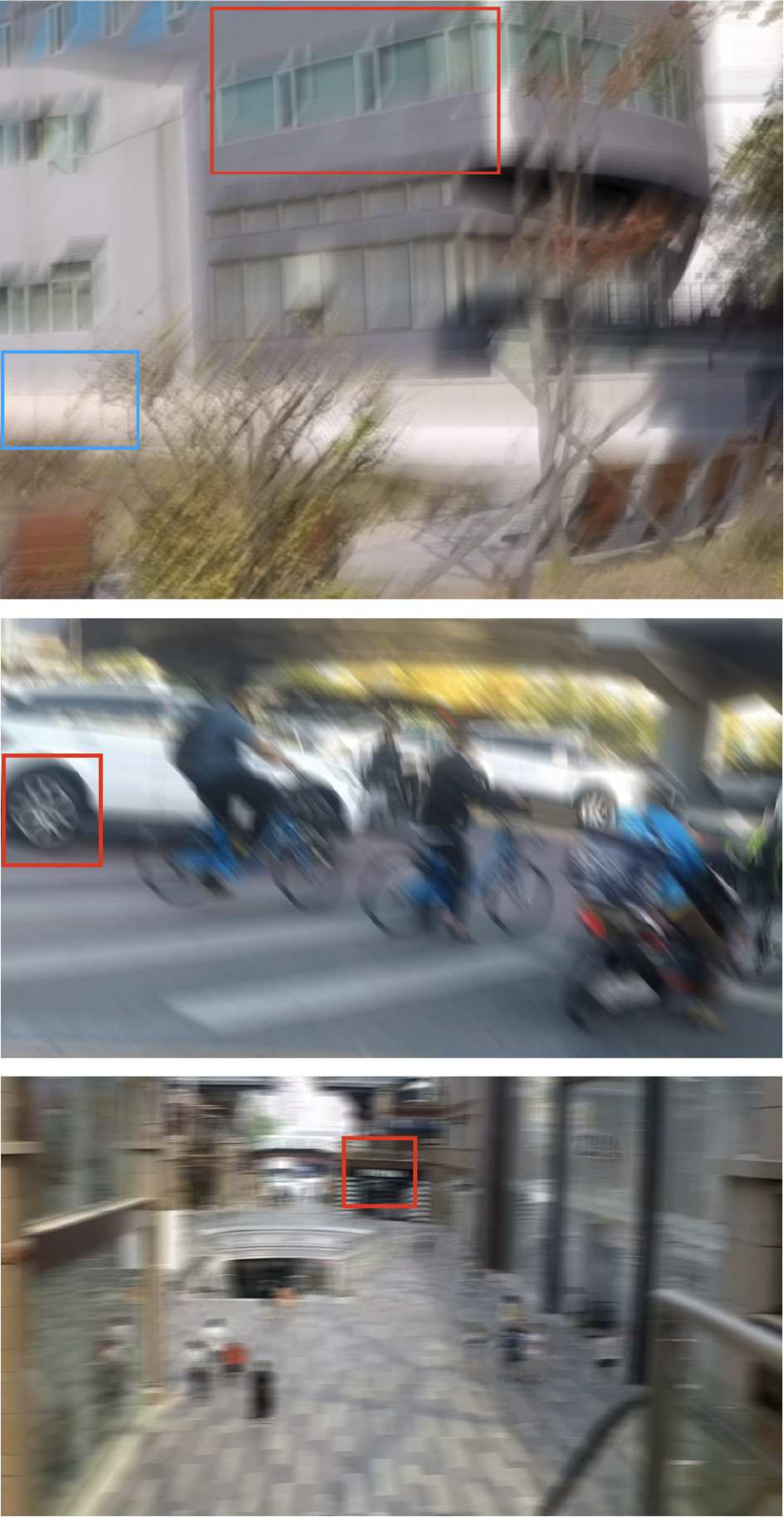}
         \caption{Blurry Image}
     \end{subfigure}
     \begin{subfigure}[b]{0.22\textwidth}
         \centering
         \includegraphics[height=0.3\textheight]{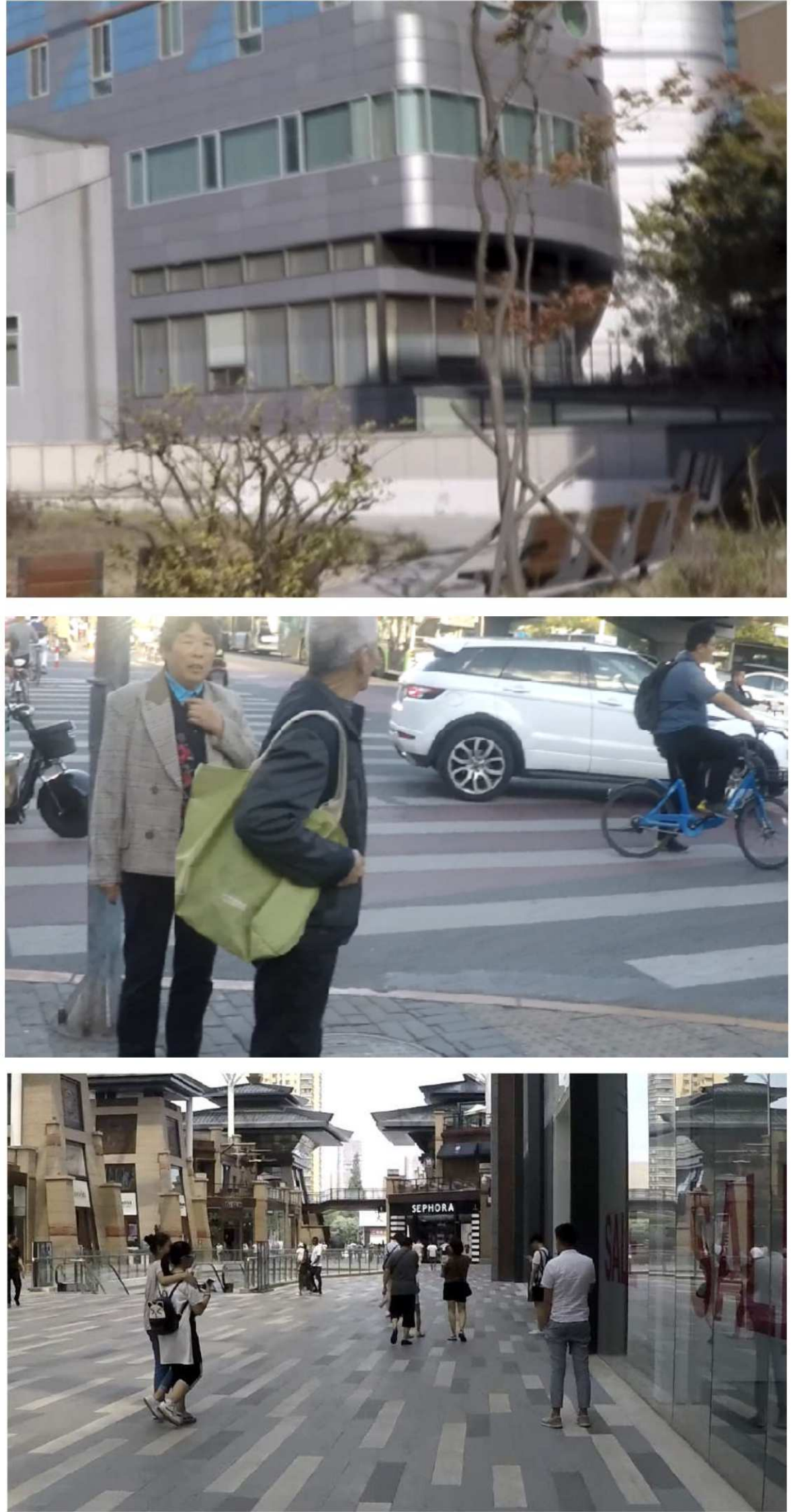}
         \caption{Ref Image}
     \end{subfigure}
     \begin{subfigure}[b]{0.13\textwidth}
         \centering
         \includegraphics[height=0.3\textheight]{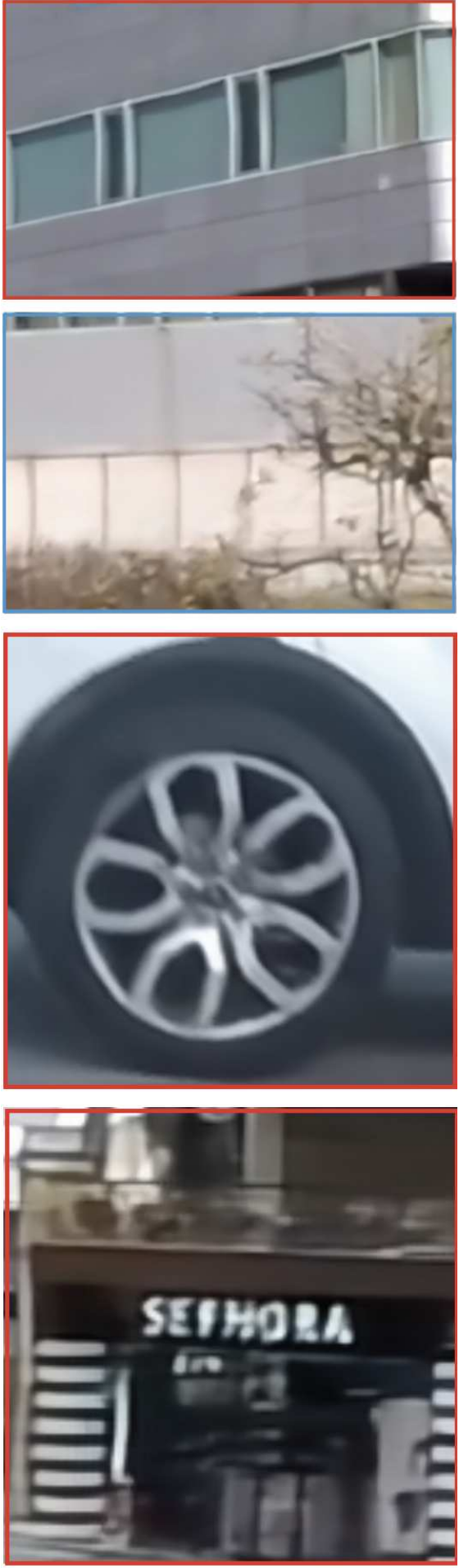}
         \caption{w/o Ref}
     \end{subfigure}
     \begin{subfigure}[b]{0.13\textwidth}
         \centering
         \includegraphics[height=0.3\textheight]{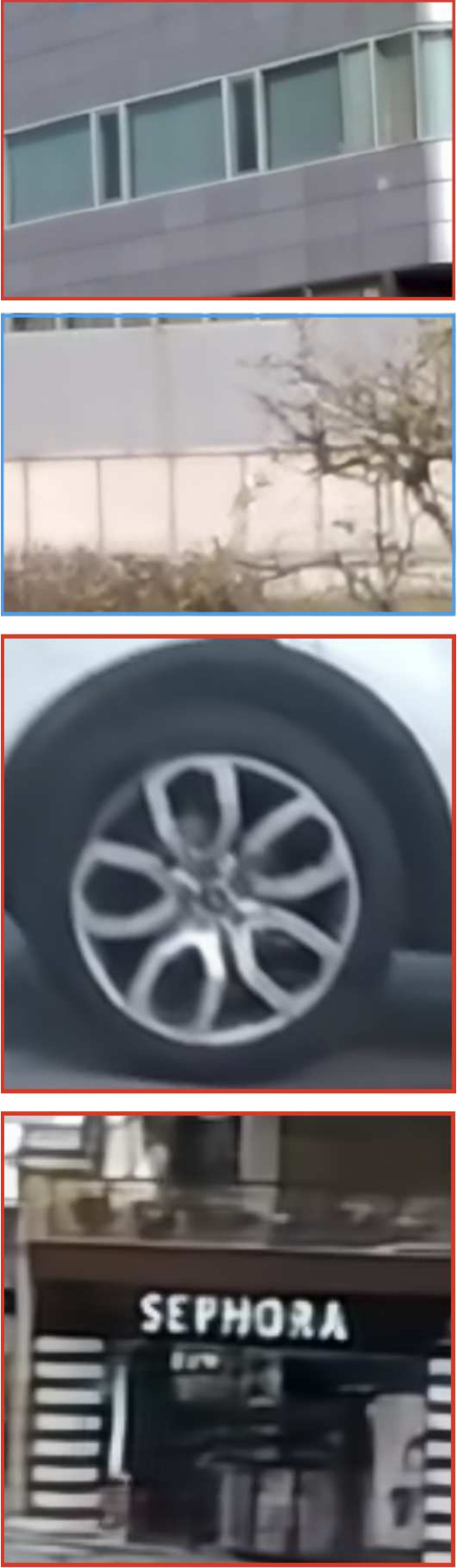}
         \caption{w Ref}
     \end{subfigure}
     \begin{subfigure}[b]{0.13\textwidth}
         \centering
         \includegraphics[height=0.3\textheight]{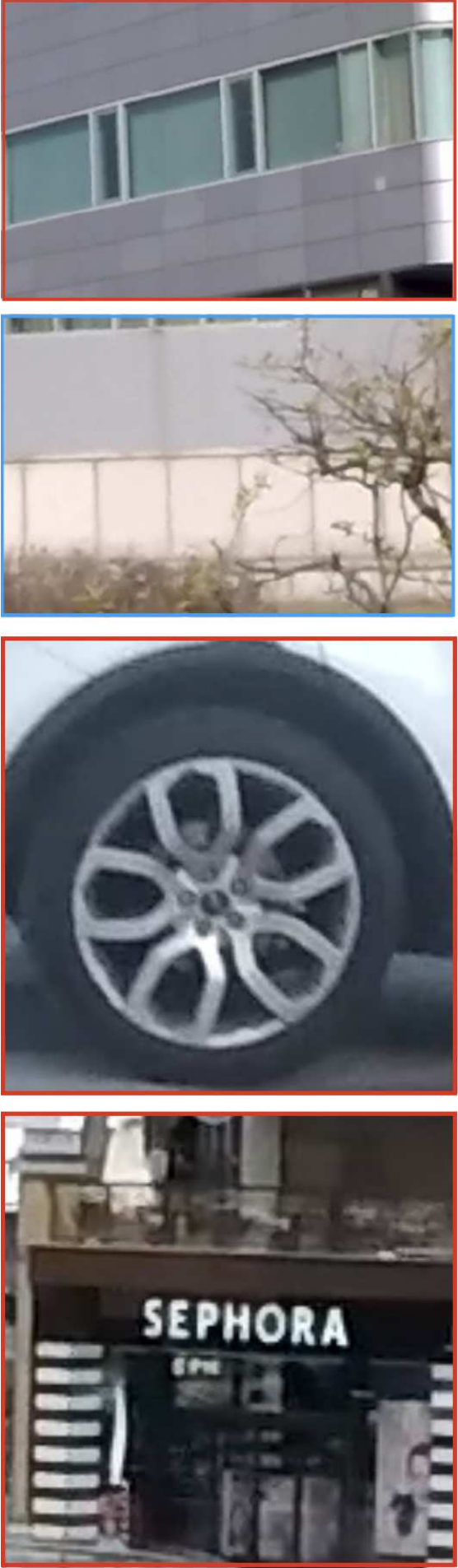}
         \caption{GT}
     \end{subfigure}
     \caption{Effects of different types of reference images on the results. }
     \label{fig:ref_comp}
\end{figure*}

The choice of a reference image will affect the result of our method. Figure \ref{fig:ref_comp} demonstrates the impact of the choice of reference images on Ref-MIMO-UNet. We select reference images with different properties here.

The image in the first row is from the GOPRO dataset. The specified reference image is chosen from the same sequence as the input; it is less blurry than the input. Column (c) shows the results obtained by the model that does not use a reference, which illustrates the upper bound performance of single image deblurring methods. Although the reference is blurry, using it as a reference leads to a better result; the edges of the windows, etc. are more straight and textures become finer. The images in the second and third rows are from HIDE. For the second image, we specify an image contains a car of the same model seen in the input image; we choose it from a different source from HIDE. We can see that the result in (d) reconstructs slightly more accurate texture of the wheel, although the reference is not so sharp. For the third image, we specify an image of the same road but from a considerably different viewpoint. Using the reference (in (d)) leads to clearly better results, such as precise reconstruction of texts. 

\subsubsection{Application to Video Deblurring}
\begin{figure}[!ht]
     \centering
     \begin{subfigure}[b]{0.19\textwidth}
         \centering
         \includegraphics[width=\textwidth]{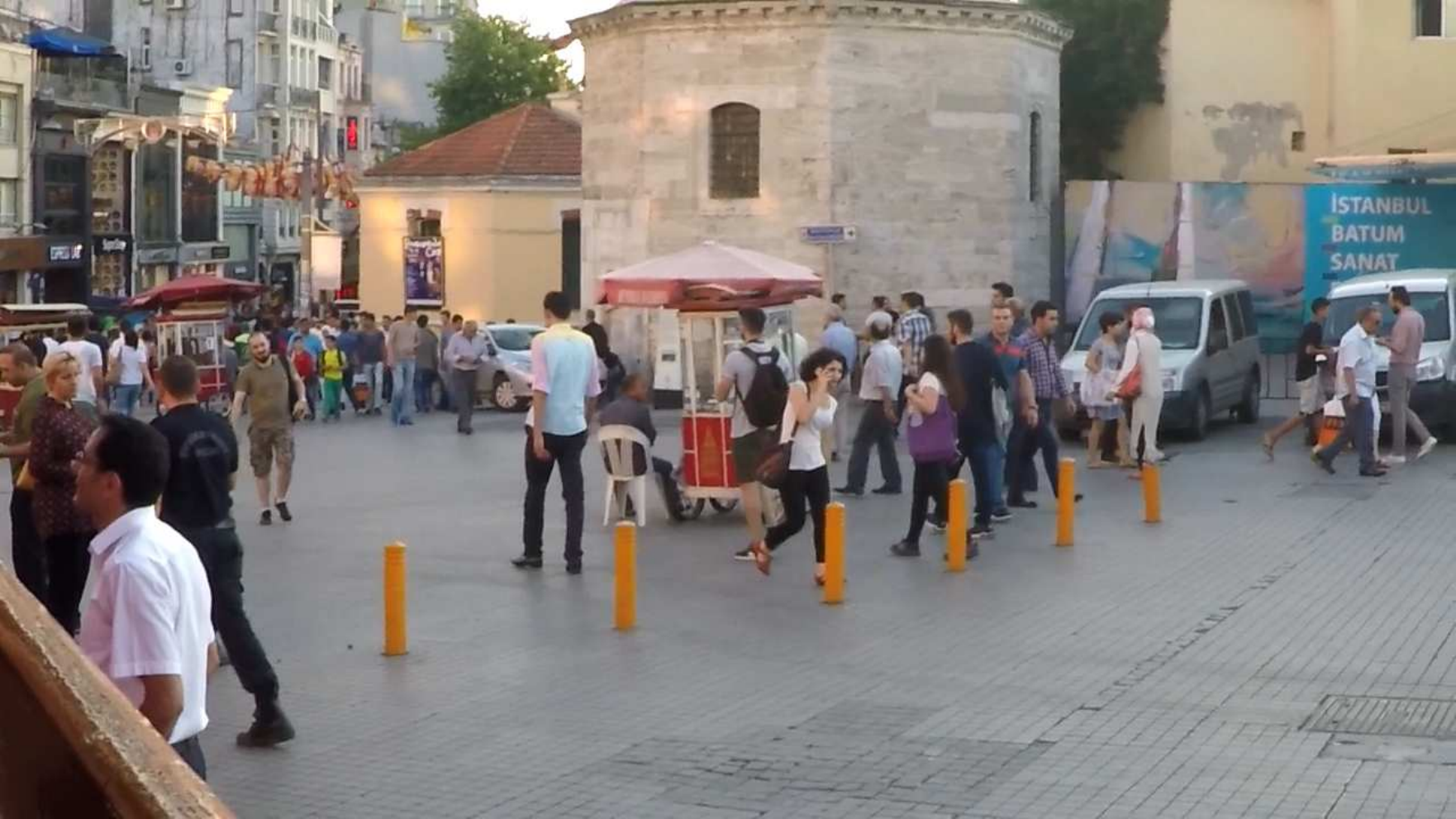}
         \caption{0.196}
     \end{subfigure}
     \begin{subfigure}[b]{0.19\textwidth}
         \centering
         \includegraphics[width=\textwidth]{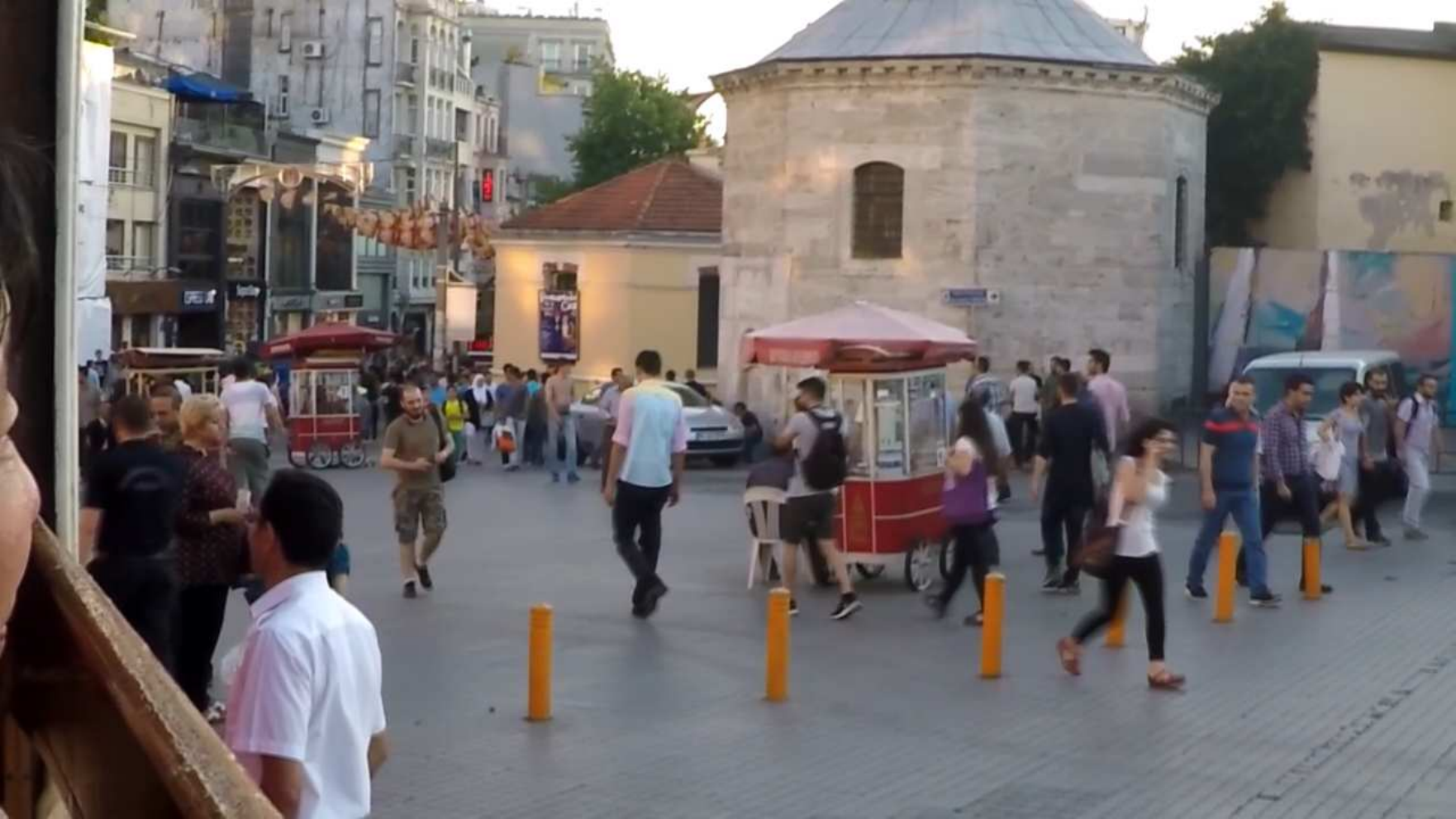}
         \caption{0.150}
     \end{subfigure}
     \begin{subfigure}[b]{0.19\textwidth}
         \centering
         \includegraphics[width=\textwidth]{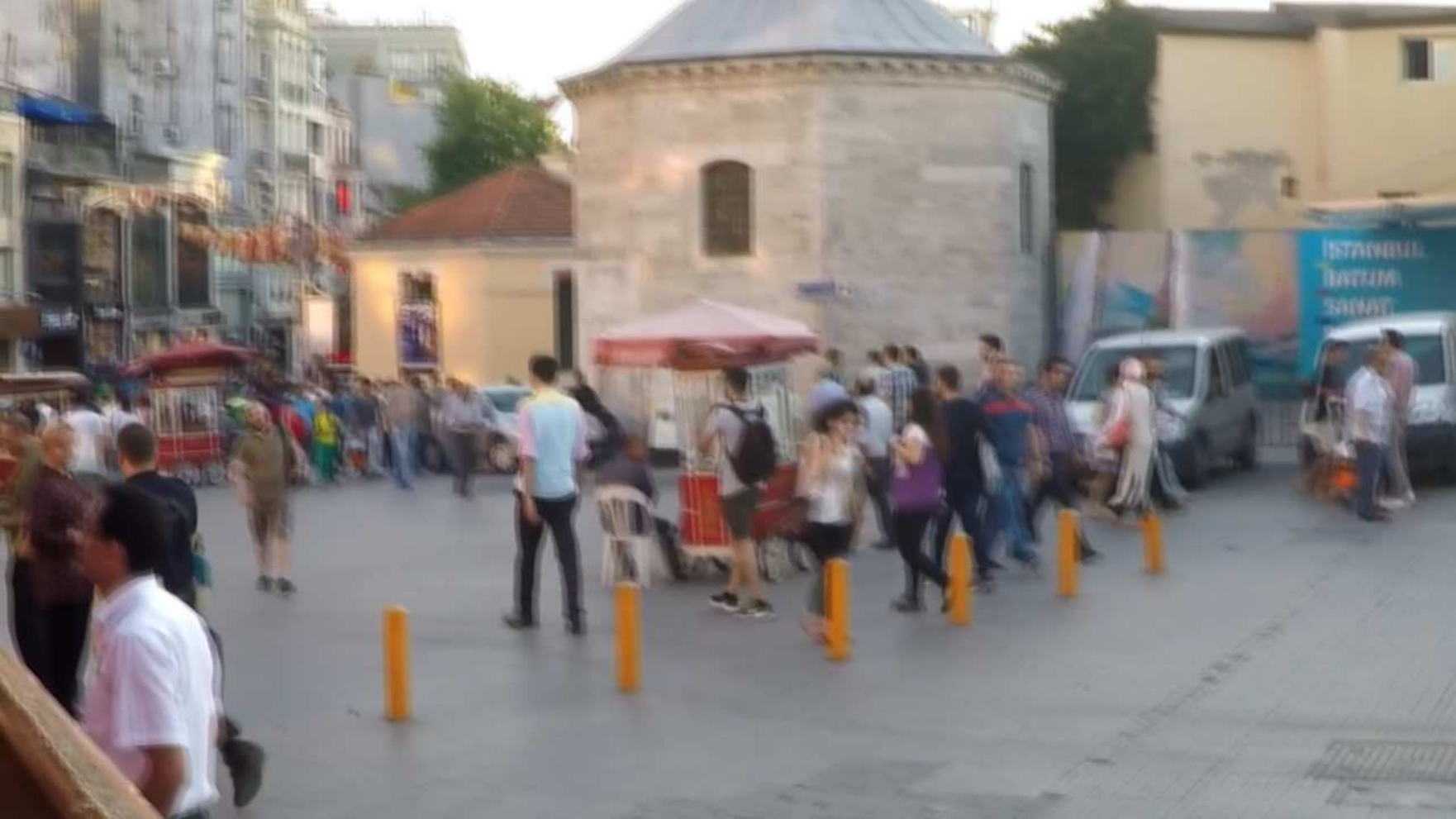}
         \caption{0.102}
     \end{subfigure}
     \begin{subfigure}[b]{0.19\textwidth}
         \centering
         \includegraphics[width=\textwidth]{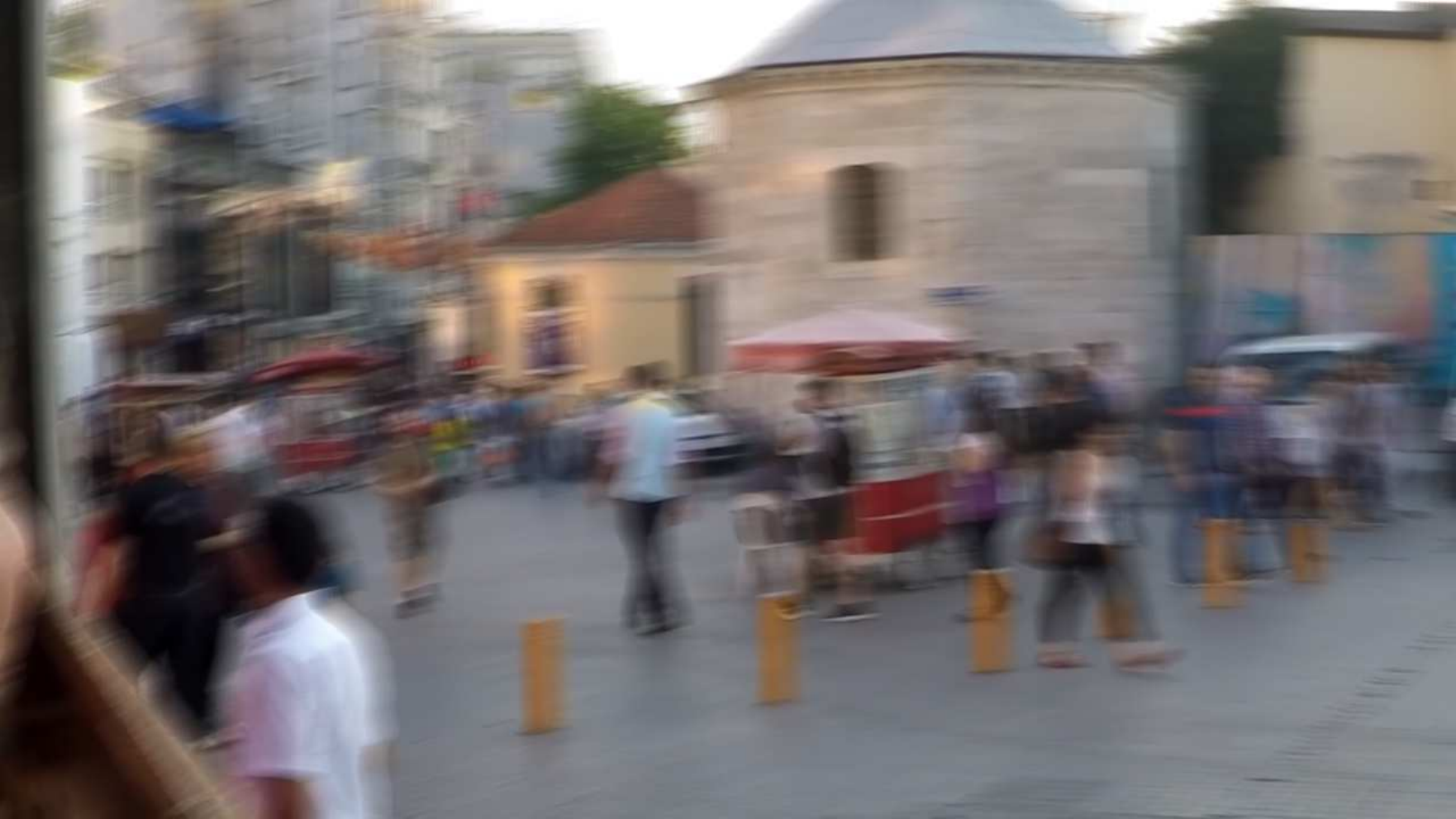}
         \caption{0.064}
     \end{subfigure}
     \caption{Relationship between blurry images and their sharpness metric. }
     \label{fig:sharpness}
\end{figure}

In real-world deblurring, we may cannot find an ideal sharp image as a reference. Considering real-world videos often consist of blurry frames with different degrees of blur. In that case, we can find sharp (or mildly blurry) frames and use them as a reference to deblur the blurry frames. To do this, we adopt a simple method for evaluating the sharpness of an image~\cite{de2013image}. The sharpness of an image $I$  with size $M\times N$ is the portion of pixels whose value is larger than $thres$,
\begin{align}
thres =  \frac {\max(|CFT(I)|)} {1000 \times M\times N},
\end{align}
where $CFT$ is the centered Fourier transform. Sharper images have higher sharpness scores.  Figure~\ref{fig:sharpness} shows several images with different level of blurs and their evaluated sharpness scores. 

 To evaluate the proposed methods on blurry Ref inputs, we conduct experiments using the GOPRO dataset, which contains images with different amounts of blur in each image sequence. Specifically, for each input, we consider the range from $-30$ to $30$ frames in the same sequence, and then choose a frame for a reference in the following four ways:
\begin{itemize}
    \item The most blurry frame
    \item The frame with intermediate blur
    \item The sharpest frame
\end{itemize}
 All the frames in the above range (60 in total) are sorted. The above three images are chosen from the top, the middle, and the bottom of the sorted list. Table~\ref{table:comp_sharp} shows the results when we specify each of the above three images for references. We can see that our method yields better results as the sharpness of a reference image increases.

\setlength{\tabcolsep}{4pt}
\begin{table*}[!htb]
\caption{Impact of the sharpness of the reference image. Results on the GOPRO dataset. See the text for details. }
\label{table:comp_sharp}
\begin{center}
\begin{tabular}{lcccccc}
\hline
\multirow{2}{*}{Methods}  &\multicolumn{2}{c}{Most blurry} & \multicolumn{2}{c}{Intermediate} &\multicolumn{2}{c}{Least blurry} \\\cline{2-7}
 & PSNR & SSIM& PSNR & SSIM& PSNR & SSIM\\
\hline
Ref-DeblurNet   & 31.39 & 0.947 & 31.45  & 0.958 & 31.56& 0.950 \\
Ref-MIMO-UNet   & 32.28 & 0.956 & 32.33 &  0.956 & 32.43 & 0.957  \\
Ref-MIMO-UNet+  & 33.01 & 0.961 & 33.04 & 0.962  & 33.11 & 0.962  \\
Ref-NAFNet32    & 33.04 & 0.962 & 33.08 & 0.962  & 33.13 & 0.963 \\
Ref-NAFNet64    & 33.96 & 0.968 & 33.99 & 0.968  & 34.05 & 0.969\\

\hline
\end{tabular}
\end{center}
\end{table*}

\subsubsection{Runtime and Params Comparison}
We compare the number of parameters and computational cost of our modified ref-net with the original models. The results are listed in Table~\ref{table:comp_resource}. The MACs column compares the computational cost of processing a $256 \times 256$ patch. 

\begin{table}[!htb]
\begin{minipage}{0.98\columnwidth}
\caption{Comparison of model parameters and GMACs to process a $256 \times 256$ image}
\label{table:comp_resource}
\begin{center}
\begin{tabular}{lcc}
\hline
Methods & Params(M)   & GMac \\\cline{2-3}
\hline
DeblurNet~\cite{zhou2019davanet}                                 & 4.6  & 37.2  \\
Ref-DeblurNet                            & 8.1   & 71.3  \\
MIMO-UNet~\cite{cho2021rethinking}   & 6.8   & 67.2 \\
Ref-MIMO-UNet                       & 10.3   & 101.3 \\
MIMO-UNet+~\cite{cho2021rethinking}  & 16.1  & 154.4 \\
Ref-MIMO-UNet+                      & 19.6  & 188.5 \\ 
NAFNet32~\cite{chen2022simple}         & 17.1 & 16.1 \\
Ref-NAFNet32                           & 20.2  & 29.0 \\
NAFNet64~\cite{chen2022simple}         & 67.9 & 63.3 \\
Ref-NAFNet64                          & 78.8  & 107.3 \\
\hline
\end{tabular}
\end{center}
\end{minipage}
\end{table}

\subsection{Summary and Conclusion}

We have proposed a new method to deblur a blurry image with the help of a reference image. While, the reference image is ideally a sharp image of the same scene as the input image, our method can utilize a less blurry image of the same scene or even a different one. The method employs a coarse-to-fine approach, in which the sharp image at each resolution is estimated and used for subsequent steps. This approach mitigates the difficulty of aligning (or rigorously matching local patches between) the blurry input image and the sharp reference image, leading to the proper fusion of their local features to recover the sharp detail of the blurry input patches. We have designed our method in the form of modules that augment state-of-the-art single-image deblurring architectures. Thus, it can theoretically be integrated into any single image deblurring method following the same coarse-to-fine method, including those to be developed in the future. The experimental results confirm the effectiveness of our approach. 
{\small
\bibliographystyle{ieee_fullname}

}

\end{document}